\documentclass[lettersize,journal]{IEEEtran}

\usepackage{amsmath,amsfonts}
\usepackage{graphicx}
\usepackage{cite}
\usepackage{url}
\usepackage{algorithmic}
\usepackage{algorithm}
\usepackage{array}
\usepackage[caption=false,font=normalsize,labelfont=sf,textfont=sf]{subfig}
\usepackage{textcomp}
\usepackage{stfloats}
\usepackage{verbatim}
\usepackage{booktabs}
\usepackage{tabularx}
\usepackage{booktabs}
\usepackage{makecell}
\usepackage{siunitx}

\usepackage{rotating}
\begin{document}

\title{MMRHP: A Miniature Mixed-Reality HIL Platform for Auditable Closed-Loop Evaluation}

% --- Author Information (Placeholder) ---
\author{Mingxin Li, Haibo Hu, Jinghuai Deng, Yuchen Xi, Xinhong Chen, Jianping Wang

\thanks{Mingxin Li, Haibo Hu, Jinghuai Deng, Yuchen Xi, Xinhong Chen and Jianping Wang are with the school of City University of Hong Kong(email: mingxin.li@cityu.edu.hk; haibohu2-c@my.cityu.edu.hk; jinghdeng2-c@my.cityu.edu.hk; yuchenxi@cityu.edu.hk; xinhong.chen@cityu.edu.hk; jianwang@cityu.edu.hk) (\emph{Mingxin Li, Haibo Hu and Jinghuai Deng contributed equally to this work})(\emph{Corresponding author: Jianping Wang.})}}

% --- Paper Headers ---
% \markboth{IEEE Transactions on Intelligent Vehicles,~Vol.~XX, No.~Y, ~2025}%
% {Author \MakeLowercase{\textit{et al.}}: MMRHP: A Miniature Mixed-Reality HIL Platform}

% \IEEEpubid{0000--0000/00\$00.00~\copyright~2025 IEEE} % Uncomment for final version
\maketitle

% --- Abstract ---
\begin{abstract}
Validation of autonomous driving systems requires a trade-off between test fidelity, cost, and scalability. While miniaturized hardware-in-the-loop (HIL) platforms have emerged as a promising solution, a systematic framework supporting rigorous quantitative analysis is generally lacking, limiting their value as scientific evaluation tools. To address this challenge, we propose MMRHP, a miniature mixed‑reality HIL platform that elevates miniaturized testing from functional demonstration to rigorous, reproducible quantitative analysis. The core contributions are threefold. First, we propose a systematic three-phase testing process oriented toward the Safety of the Intended Functionality (SOTIF) standard, providing actionable guidance for identifying the performance limits and triggering conditions of otherwise correctly functioning systems. Second, we design and implement a HIL platform centered around a unified spatiotemporal measurement core to support this process, ensuring consistent and traceable quantification of physical motion and system timing. Finally, we demonstrate the effectiveness of this solution through comprehensive experiments. The platform itself was first validated, achieving a spatial accuracy of 10.27 mm RMSE and a stable closed-loop latency baseline of approximately 45 ms. Subsequently, an in-depth Autoware case study leveraged this validated platform to quantify its performance baseline and identify a critical performance cliff at an injected latency of 40 ms. This work shows that a structured process, combined with a platform offering a unified spatio-temporal benchmark, enables reproducible, interpretable, and quantitative closed‑loop evaluation of autonomous driving systems.
\end{abstract}

% --- Keywords ---
\begin{IEEEkeywords}
Autonomous Driving, Hardware-in-the-Loop (HIL), Mixed Reality, CARLA, SOTIF, Validation and Verification (V\&V).
\end{IEEEkeywords}

\section{Introduction}\label{sec:intro}

\IEEEPARstart{T}{he} commercial deployment of autonomous vehicles (AVs) faces a critical bottleneck that has shifted from achieving basic functionality to delivering statistically convincing safety in long-tail scenarios~\cite{KALRA2016182}. The risks in these long-tail cases often do not stem from conventional hardware or software failures, but from performance limitations of otherwise correctly functioning systems under specific conditions. To systematically address this, the Safety Of The Intended Functionality (SOTIF, ISO 21448) standard emerged, extending the traditional notion of functional safety to explicitly cover safety risks arising from system performance constraints~\cite{development_ISO_21448}. As such, SOTIF practice is not merely about increasing the number of test scenarios. Its core requirements are identifying triggering conditions and iteratively validating risk reduction, as also advocated by the industry practices~\cite{pegasus2019method,WOS_000982607400012}.

In pursuing SOTIF’s scientific measurement paradigm, the Verification \& Validation (V\&V) community encounters two interrelated problems~\cite{birkemeyer2023scenario}. The first widely studied one is the \emph{What to Test} problem, which concerns how to systematically generate or select critical scenarios that effectively probe system performance boundaries~\cite{hou2023twin,riedmaier2020survey}. To execute these scenarios, the V\&V community employs a wide spectrum of testing platforms, ranging from purely virtual simulations to physical testbeds~\cite{pietruch2020overview, zhang2024virtual}. 
The second one, more fundamental yet underexplored, is the \emph{How to Measure} problem~\cite{rajabli_svvsac_review,birkemeyer2023scenario}: even with a perfect scenario suite, how do we quantify behavior and performance with scientific rigor, reproducibility, and auditability?
The absence of a unified, general measurement framework for closed-loop evaluation constitutes what we define as the \emph{Measurement Dilemma}. To systematically address both problems, we argue for a validation platform that can support large-scale scenario testing while maintaining physical fidelity for a rigorous measurement system.

To this end, the first challenge is a trilemma rooted in the What to Test problem: it is difficult to simultaneously optimize Physical Fidelity, Scalability \& Efficiency, and Cost. Pure Software-in-the-Loop (SiL) simulation offers high scalability and low cost, but the inherent sim-to-real gap raises concerns about physical fidelity~\cite{zhang2024survey}. Conversely, full-scale Vehicle-in-the-Loop (ViL) road testing achieves the highest fidelity but at prohibitive cost and very low efficiency, making it unsuitable for the large-scale scenario validation SOTIF demands~\cite{KALRA2016182,liu2024vil}. Early efforts to bridge this SiL-ViL gap pioneered mixed-reality platforms~\cite{quinlan2010bringing, gechter2014towards}, such as the Mcity testbed, where vehicles interact with dense virtual traffic to evaluate safety-critical scenarios~\cite{feng2018augmented}.Despite the wide adoption of this approach~\cite{varga2020mixed,kneissl2020mixed,szalai2020mixed,drechsler2022dynamic}, they remain fundamentally constrained by the high operational costs and low scalability. Against this backdrop, a more scalable and cost-effective compromise has emerged: mixed-reality miniature testbeds that combine Hardware-in-the-Loop (HIL)~\cite{mokhtarian2024survey,li2025autonomous,Tae_sil_mr_agv_ap}. They significantly lower development barriers and cost by enabling rapid, safe, and low-cost iteration on hardware without the risks and cost of full-scale vehicle testing~\cite{verma2021implementation, vargas2024design, balaji2019deepracer, bulsara2020obstacle, argui2023mixed-reality, liu2020mobile, chen2009mixed}. 

% --- FIGURE fig_sys_panoramic_view: panoramic view ---
\begin{figure*}[!t]
    \centering
    \includegraphics[width=\textwidth]{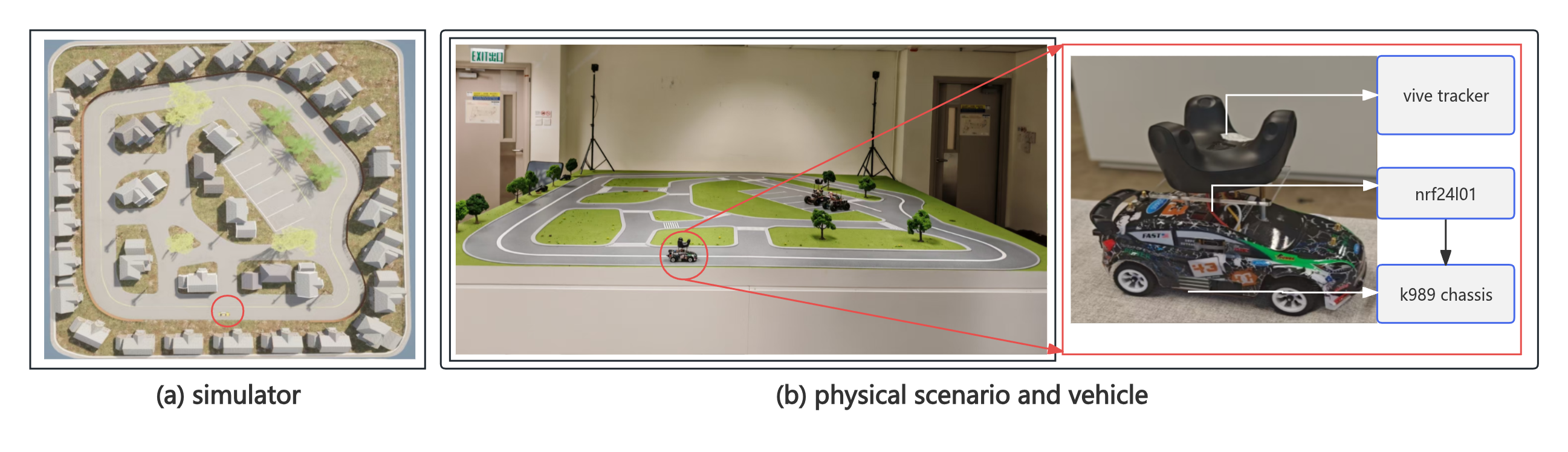}
    \caption{An overview of the MMRHP platform.(a) The virtual town scene rendered in the CARLA simulator. (b) The 4.2m x 4.2m real-world, scaled-down sandbox environment, which includes various road elements and a scaled-physical vehicle.}
    \label{fig:panoramic}
\end{figure*}

While these miniature mixed-reality hardware-in-the-loop (mini-MR-HIL) platforms offer a better balance in the What to Test trilemma, a closer examination reveals they do not convincingly address the How to Measure dilemma. This metrological gap appears in different forms across the main platform archetypes. Platforms like UDSSC~\cite{stager2018scaled} provide high-precision motion capture data, creating an ideal environment for validating control algorithms. However, this design bypasses the SUT's perception stack, thus overlooking the potential impact of perceptual uncertainties and precluding a true end-to-end performance evaluation. Furthermore, it lacks a unified temporal reference frame for diagnostics, which is essential for synchronizing events across physical and virtual domains. Subsequently, platforms like CPM Lab~\cite{kloock2021cyber} introduce a synchronization middleware, but still omit the full AD stack. More significantly, its synchronization mechanism is engineered for internal coordination and is not intended to serve as a stable reference for validating system performance. Finally, platforms like ICAT~\cite{tian2024icat} pursue a more realistic simulation of decentralized systems by forgoing motion capture. While this choice improves fidelity, it also discards the high-value evaluation metric of external ground-truth reference. This trade-off presents a significant challenge, as the confounding of algorithmic faults with the SUT's own perceptual uncertainties makes it difficult to generate auditable data.

These recurring analytical roadblocks motivate a clear need for a new platform \textbf{fundamentally designed} from the ground up to close this metrological gap. To effectively address this, a measurement-first solution must meet three core requirements. First, to trace data flow through the entire system, it must guarantee \emph{auditable data}. Second, to precisely synchronize events across different subsystems, it must establish \emph{unified timing}. Third, to isolate sources of performance-degrading latency, it must provide a \emph{decomposable latency model}. We contend that these three features are the foundational components of any \textbf{effective solution} to the How to Measure dilemma.

To meet these needs and close the metrological gap, we design MMRHP, a solution aimed at elevating miniature HIL testing from functional demonstration to rigorous scientific experimentation. Our approach comprises two core components. First, to provide a concrete methodology for implementing the SOTIF standard, we propose a three-stage workflow. This workflow operationalizes the SOTIF process by systematically moving from \textbf{establishing a performance baseline} in known-safe conditions (Stage 1), to \textbf{identifying and quantifying the impact of specific triggering conditions} through controlled perturbation (Stage 2), and finally to \textbf{evaluating system robustness} in complex scenarios composed of multiple interacting triggers (Stage 3). This structured process provides a clear path to delineating the system's performance boundaries as required by SOTIF. Second, we implement a \emph{measurement-first HIL platform} to support this workflow. The platform provides a scalable town environment and features an independent \emph{measurement core}. This core consists of an external ground truth system, a unified clock domain, and a decomposable latency model. Together, these components ensure that all scenario tests yield reproducible and auditable results, thereby mitigating the measurement dilemma and enabling attributable error diagnosis.

The main contributions are as follows:
\begin{enumerate}
\item \textbf{A systematic, SOTIF-compliant evaluation workflow} for quantifying the performance boundaries of an autonomous system.
\item \textbf{A measurement-first MR-HIL mini platform} that enables auditable, closed-loop scenario testing with high-fidelity reproduction.
\item \textbf{Comprehensive validation achieving \textbf{10.27,mm} RMS spatial error and a \textbf{45,ms} closed-loop latency baseline}, alongside a case study demonstrating the workflow's effectiveness on a Autonomous Driving stack.
\end{enumerate}

The remainder of this paper details the system's design, validation, and application, demonstrating a practical path toward credible, audit-ready closed-loop evaluation.

\section{Related Work}
\label{sec:related_work}

The verification and validation (V\&V) landscape for autonomous driving has diversified into complementary directions with distinct trade-offs among physical fidelity, scalability/efficiency, and metrological rigor~\cite{li2025autonomous,mokhtarian2024survey}. We group representative works into three categories to position our contribution and motivate a measurement-first paradigm.

%=============table relatedwork========
\begin{table*}[!htbp]
    \centering
    \caption{Comparison of Related Works and Positioning of This Work (MMRHP, Min. $=$ Miniature, Ful. $=$ Full-Scale, Ex. $=$ External, On. $=$ Onboard)}
    \label{tab:platform_comparison}
    \renewcommand{\arraystretch}{1} 
    \begin{tabularx}{\textwidth}{>{\raggedright\arraybackslash}p{1.8cm} >{\raggedright\arraybackslash}p{1.7cm} >
    {\raggedright\arraybackslash}p{2.8cm} >{\raggedright\arraybackslash}p{4cm} >{\raggedright\arraybackslash}p{6cm}}
        \toprule
        \textbf{Platform / Framework} & \textbf{Scale} & \textbf{Localization Scheme} & \textbf{Primary Research Goal} & \textbf{Key Contribution / Method} \\
        \midrule

        UDSSC~\cite{stager2018scaled} & Min. (1:25) & Ex. (Vicon) & Multi-Agent Coordination &  traffic flow optimization. \\
        \addlinespace

        Cambridge~\cite{hyldmar2019fleet} & Min. (1:24) & Ex. (OptiTrack) & Multi-Agent Coordination &  testing classic driver models (IDM/MOBIL) . \\
        \addlinespace

        CPM Lab~\cite{kloock2021cyber} & Min. (1:18) & Ex.(Single Camera) & Distributed Control &  deterministic, reproducible distributed experiments. \\
        \midrule

        ICAT~\cite{tian2024icat} & Min. (Custom) & On.(LiDAR SLAM) & End-to-End System Replication & Full software stack testing. \\
        \addlinespace

        MCCT~\cite{dong2023mixed} & Min. (1:14) & Ex.(Roadside Cam) & Mixed-Reality Interaction & ``mixedDT'' in cloud-controlled  \\
        \midrule

        HIL+FI~\cite{abboush2024virtual} & Ful. (ECU) & N/A (HIL) & Functional Safety & Non-invasive, real-time fault injection at the CAN bus level without modifying system models. \\
        \addlinespace
        SciL~\cite{szalay2021next} & Ful. (Vehicle) & Ex.(RTK-GNSS) & Scenario-Based Validation & Controls the entire mixed-reality scene, not just stimulating a single vehicle. \\
        \midrule
        \textbf{Our Work MMRHP} & \textbf{Min.} & \textbf{Ex.} & \textbf{Algorithm Performance Diagnosis} & \textbf{Built-in, auditable metrology and active perturbation injection for reproducible scientific experiments.} \\
        \bottomrule
    \end{tabularx}
\end{table*}

\subsection{Miniature Platforms for Multi-Agent Traffic}
Miniature testbeds with high-precision external localization supply global ground truth, suppressing perception/self-localization uncertainty and enabling reproducible studies of coordination and macroscopic traffic effects. UDSSC uses Vicon motion capture, a centralized main frame, and 'virtual reference + reference tracking' control, demonstrating stop and go elimination, a reduction in travel time of 18.7\% and improved energy use in a ten-vehicle merging experiment~\cite{stager2018scaled}. Related platforms advance accessibility and determinism: Cambridge Minicar validates IDM/MOBIL and quantifies V2V throughput gains~\cite{hyldmar2019fleet}; CPM Lab contributes DDS middleware with LET synchronization for deterministic, reproducible distributed control and human-in-the-loop studies~\cite{kloock2021cyber,scheffe2023scaled}. These platforms excel at macro-phenomena and distributed control but, by replacing onboard perception/localization with external ground truth and often external coordination, are not tailored for metrological diagnosis of single-agent performance boundaries under realistic sensing/processing uncertainty.

\subsection{High-Fidelity End-to-End Platforms for System Integration}
A second direction replicates the full perception--planning--control loop and compute stack at miniature scale for integrated, closed-loop evaluation. ICAT runs complete stacks (e.g., Autoware/SOAFEE) on Jetson, uses 2D LiDAR with NDT-SLAM for decentralized localization, and bridges CARLA/SUMO twins with an indoor V2X testbed~\cite{tian2024icat}. MCCT’s mixedDT fuses physical miniature vehicles, virtual agents, and roadside perception in a cloud-orchestrated mixed reality~\cite{dong2023mixed}. These platforms are strong for end-to-end performance, yet attribution suffers from \emph{error coupling}: degraded behavior may stem from algorithmic issues or upstream noise, latency, and tracking errors.  ICAT reports such phenomena (initial pose sensitivity, curved-path tracking error, and lag-induced response changes)~\cite{tian2024icat}. Furthermore, their reliance on perception within feature-sparse indoor environments raises critical questions about the generalization of findings to real-world scenarios. This motivates the need for independent ground truth, a unified clock domain, and a decomposable latency model for auditable attribution.

\subsection{System-Level Frameworks for V\&V and Safety}
A third category of work focuses on system-level frameworks spanning the V-model, aiming to streamline validation through co-simulation, fault injection, and scene orchestration. Their primary focus is on \textbf{development efficiency and functional safety}. Consequently, their reported metrics often center on system-level outcomes or communication health (e.g., pass/fail rates, round-trip times~\cite{szalay2021next}), rather than on providing a \textbf{decomposable, auditable measurement chain} to explain \textit{why} a subtle performance degradation occurred. This emphasis is evident across the literature: some frameworks enable developers to ``code once, test anywhere''  for portability, but without a focus on metrological diagnosis~\cite{bruggner2021model}; others concentrate on functional safety via real-time fault injection~\cite{abboush2024virtual}; and still others, like ZalaZONE's SciL, excel at orchestrating complex scenes~\cite{szalay2021next}.

\subsection{Summary and Positioning}
In summary, existing platforms respectively target multi-agent coordination, end-to-end integration, or system-level safety orchestration. Our work addresses the under-explored gap for a platform that \textbf{centers auditable, reproducible measurement} to diagnose algorithmic performance boundaries. The MMRHP is designed for this niche: it integrates the external ground truth and reproducibility of coordination testbeds to avoid the error coupling pitfalls of end-to-end platforms, while providing the embedded metrology core that system-level frameworks lack. By doing so, it offers a practical and rigorous pathway toward SOTIF-compliant evaluation.

\section{System Design: A HIL Platform for Reproducible Performance Evaluation}
\label{sec:system_design}
This section details our complete solution to the Measurement Dilemma in autonomous driving systems, which forms the technical core of the paper. The solution consists of two parts. First, we propose a principled workflow for systematic performance analysis. Second, we introduce the supporting technical implementation, the MMRHP, a miniature HIL platform focused on verifiable and traceable measurement capabilities.

The structure of this chapter follows a logical progression from analysis workflow to platform implementation:
\begin{itemize}
    \item \textbf{Section \ref{sec:workflow}} will first define a three-stage workflow aimed at transforming the evaluation of autonomous driving systems from functional demonstrations to systematic performance analysis.
    \item \textbf{Sections \ref{sec:measurement_framework} through \ref{sec:arch_and_impl}}will provide a detailed introduction to the core measurement framework (Sec. \ref{sec:measurement_framework}) that supports this workflow, as well as the hardware-in-loop collaborative driving test system (Sec. \ref{sec:arch_and_impl}) designed based on this framework and its specific engineering implementation.
\end{itemize}

\subsection{A Three-Stage Workflow for Performance Analysis}
\label{sec:workflow}
Based on the analysis in the sec.~\ref{sec:intro}, we propose a structured, three-stage methodology to systematically characterize the system's performance boundaries from different perspectives.The workflow begins by defining a stable baseline (Stage 1), and then proceeds to probe the system's limits through two complementary analysis dimensions: sensitivity to internal system parameters (Stage 2) and performance quantification under external environmental complexity (Stage 3).

\subsubsection{Stage 1: Defining the Performance Baseline}
\label{sec:workflow_stage1}
The objective of this stage is to establish the SUT's nominal performance profile, defining the ``known-safe'' operational domain. This profile serves as the essential reference against which performance boundaries will be measured. The implementation relies on the orchestration layer guiding the SUT along a predefined reference path $\Gamma$ under ideal, interference-free conditions. During execution, the high-precision GTS continuously captures the vehicle's physical trajectory $X$, enabling the platform to automatically compute and record baseline KPIs, primarily the spatial \textbf{Cross-Track Error (CTE)} and the SUT's internal processing latency (\textbf{$\Delta T_{\text{SUT}}$}).

\subsubsection{Stage 2: Mapping Boundaries via Internal Parameter Sensitivity}
\label{sec:workflow_stage2}
This stage aims to map the SUT's performance boundary against specific, isolated \textbf{internal or systemic triggering conditions}. This is achieved using the \textbf{programmable perturbation injector} (Sec.~\ref{sec:l_hil_core_loop}), which systematically introduces controlled disturbances. For instance, to map the boundary against communication delay, a series of fixed latencies (e.g., \texttt{\{0, 10, ..., 80\}}~ms) is injected into the Virtual-Real Synchronous Link. By repeating the baseline task for each configuration, the system generates a \textbf{performance response curve} (e.g., CTE vs. added latency). This curve explicitly reveals the boundary where system performance degrades due to a specific systemic stressor.

\subsubsection{Stage 3: Mapping Boundaries via External Scenario Robustness}
\label{sec:workflow_stage3}
Complementary to Stage 2, this stage aims to identify performance boundaries that emerge from complex \textbf{external or environmental triggering conditions}. Instead of varying a single internal parameter, this stage stress-tests the SUT by placing it in dynamic, multi-agent scenarios. The implementation leverages a \textbf{scripted scenario injection mechanism}, activating complex traffic flows (e.g.,five NPC vehicles of varying sizes arrive at an intersection at the same time and need to alternately pass through.) when the SUT enters a trigger zone. The goal is to determine if the system can maintain safe operation (measured by \textbf{TTC} and \textbf{$D_{\min}$}) when faced with high-density, unpredictable interactions. The \textbf{auditable data interface} logs all relevant variables, enabling \textbf{event-centric root cause analysis} for any observed safety violations, thereby revealing performance boundaries under complex environmental pressure.

% ------------------------- Figure: HIL Timing Diagram -------------------
\begin{figure*}[!t]
\centering
\includegraphics[width=\textwidth]{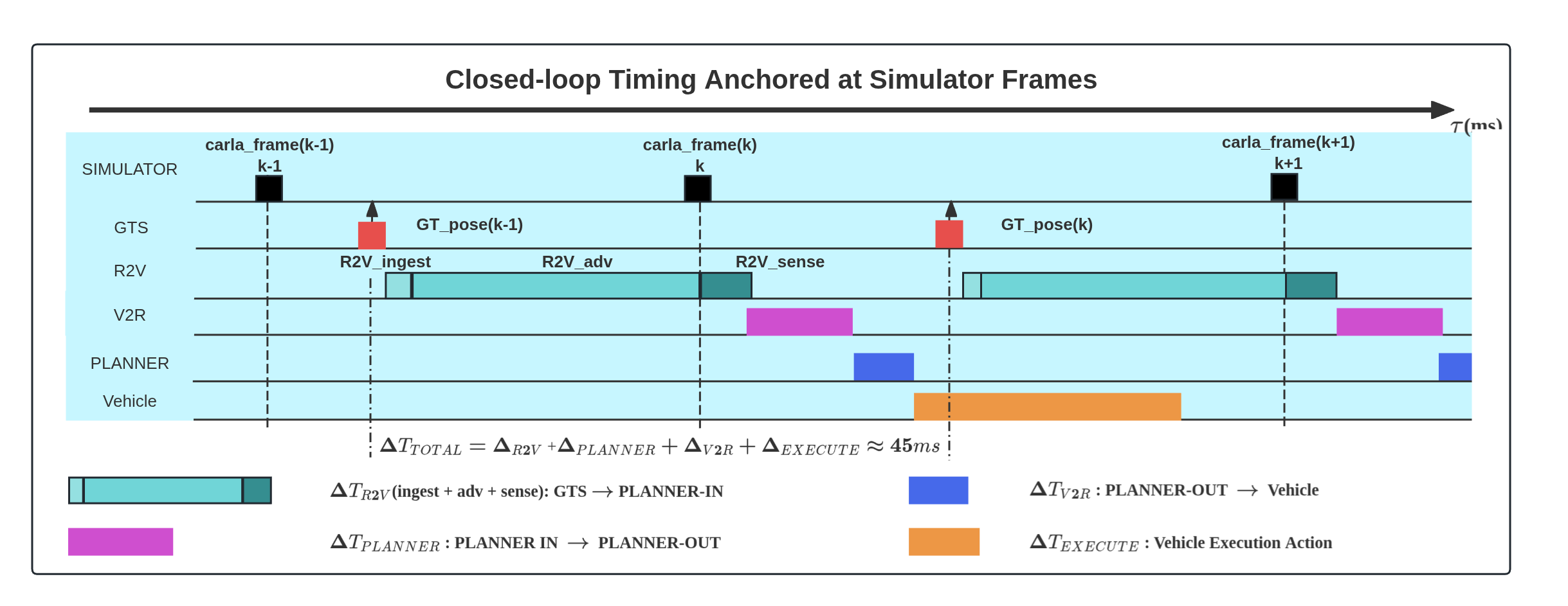}
\caption{HIL Loop Timing Diagram. This diagram illustrates the complete timing chain from ground truth acquisition to the R2V link, and then to the V2R link transmitting control commands to the physical actuator, with the measurement aperture for each latency segment $\Delta T$ indicated.}
\label{fig:timing_diagram}
\end{figure*}

\subsection{Measurement Framework: Enabling Verifiable and Traceable Evaluation}
\label{sec:measurement_framework}
To scientifically execute the workflow defined in Section~\ref{sec:workflow}, a framework that is independent of the SUT and whose measurement process is itself verifiable is indispensable. This section details the proposed measurement framework, which ensures the internal consistency and data traceability for all spatial and temporal metrics.

\subsubsection{Spatial Metrology Baseline: Reference Path Error Model}
\label{sec:path_error}
To achieve a fair comparison between different SUTs, we adopt a unified spatial error measurement method. This method does not depend on the internal state of the SUT but focuses solely on its final physical performance. We define a geometrically ideal path, the Reference Path ($\Gamma$), as the objective baseline for all measurements. The vehicle's actual physical trajectory, recorded in real-time by the external ground truth system, is termed the Actual Trajectory ($X$).

For any point $p \in X$ on the actual trajectory, we first find its projection point $p^*$ on the reference path $\Gamma$, which is the point on $\Gamma$ closest to $p$:
\begin{equation}
\label{eq:projection}
p^* = \arg\min_{q \in \Gamma} \| p - q \|_2 
\end{equation}
Based on this, the \textbf{Cross-Track Error (CTE)} is defined as the Euclidean distance from point $p$ to its projection point $p^*$:
\begin{equation}
\label{eq:cte}
\text{CTE}(p) = \| p - p^* \|_2 
\end{equation}
The Along-Track Error (ATE) measures the deviation of the vehicle in the direction of path progression. If we let $s(q)$ be the arc length of point $q$ on its trajectory and $s_d(t)$ be the desired arc length at time $t$, then ATE can be defined as:
\begin{equation}
\label{eq:ate}
\text{ATE}(p(t)) = s(p^*) - s_d(t) 
\end{equation}
This set of definitions \eqref{eq:projection}-\eqref{eq:ate} rigorously decomposes the complex spatial performance evaluation into a clear quantification of two orthogonal components. ATE is most suitable for sub‑scenarios with near‑constant velocity, low curvature, and low acceleration. Therefore, \textbf{we primarily report CTE, which is more robust across conditions.} 

\subsubsection{Temporal Metrology Baseline: Black-Box Latency Decomposition}
\label{sec:latency_decomposition}
In the time domain, we treat the SUT as a black box, measuring only its externally observable latency characteristics. To ensure measurement self-consistency, all event timestamps are uniformly recorded in a single monotonic clock domain $\tau$ (sourced from \verb|time.monotonic_ns()|).

We decompose the end-to-end total latency $\Delta T_{\text{total}}$ into two main, non-overlapping components: the SUT latency (\textbf{$\Delta T_{\text{SUT}}$}) and the inherent platform latency ($\Delta T_{\text{platform}}$). The platform latency is further subdivided into Virtual-to-Real latency ($\Delta T_{\text{V2R}}$) and Real-to-Virtual latency ($\Delta T_{\text{R2V}}$). To provide deeper analytical capabilities, we further decompose $\Delta T_{\text{R2V}}$ into $\Delta T_{\text{ingest}}$ (processing and queuing time), $\Delta T_{\text{adv}}$ (simulation advancement time), and $\Delta T_{\text{sense}}$ (virtual sensing time). This structured decomposition allows us to precisely attribute the source of latency, which is fundamental to the causal analysis defined in Section~\ref{sec:workflow}.

Ultimately, the total system latency $\Delta T_{\text{total}}$ can be modeled as a linear superposition of the segment latencies:
\begin{equation}
\label{eq:latency_total}
\Delta T_{\text{total}} = \Delta T_{\text{SUT}} + \Delta T_{\text{platform}} 
\end{equation}

where the inherent platform latency $\Delta T_{\text{platform}}$ is in turn composed of the two asymmetric data links:
\begin{equation}
\Delta T_{\text{platform}} = \Delta T_{\text{V2R}} + \Delta T_{\text{R2V}} 
\end{equation}
To enable the fine-grained diagnostics illustrated in Fig.~\ref{fig:timing_diagram}, the R2V link latency is broken down into its constituent stages:
\begin{equation}
\Delta T_{\text{R2V}} = \Delta T_{\text{ingest}} + \Delta T_{\text{adv}} + \Delta T_{\text{sense}} 
\end{equation}
This hierarchical model provides a complete framework for latency attribution. The empirical characterization of each component is detailed in the platform validation section\ref{sec:valid_b_e_link}, where we also assess the model's completeness for each link.

\subsubsection{Interaction Safety Metrics}
\label{sec:safety_metrics}
For the complex scenarios in Stage 3, the framework's metrics extend beyond the ego-vehicle's path error. It integrates classic safety metrics computed from the synchronized states of all agents in the scene, primarily including the following:
\begin{itemize}
    \item \textbf{Time-to-Collision (TTC)}: the time for two agents to collide if they maintain their current state of motion.
    \item \textbf{Minimum Inter-Agent Distance ($D_{\min}$)}: the minimum Euclidean distance between the ego-vehicle and other agents throughout the entire interaction.
\end{itemize}
These continuous metrics support dynamic risk assessment throughout the interaction process.

% --- FIGURE fig_system_architecture: MAIN SYSTEM ARCHITECTURE DIAGRAM ------
\begin{figure*}[!t]
    \centering
    \includegraphics[width=\textwidth]{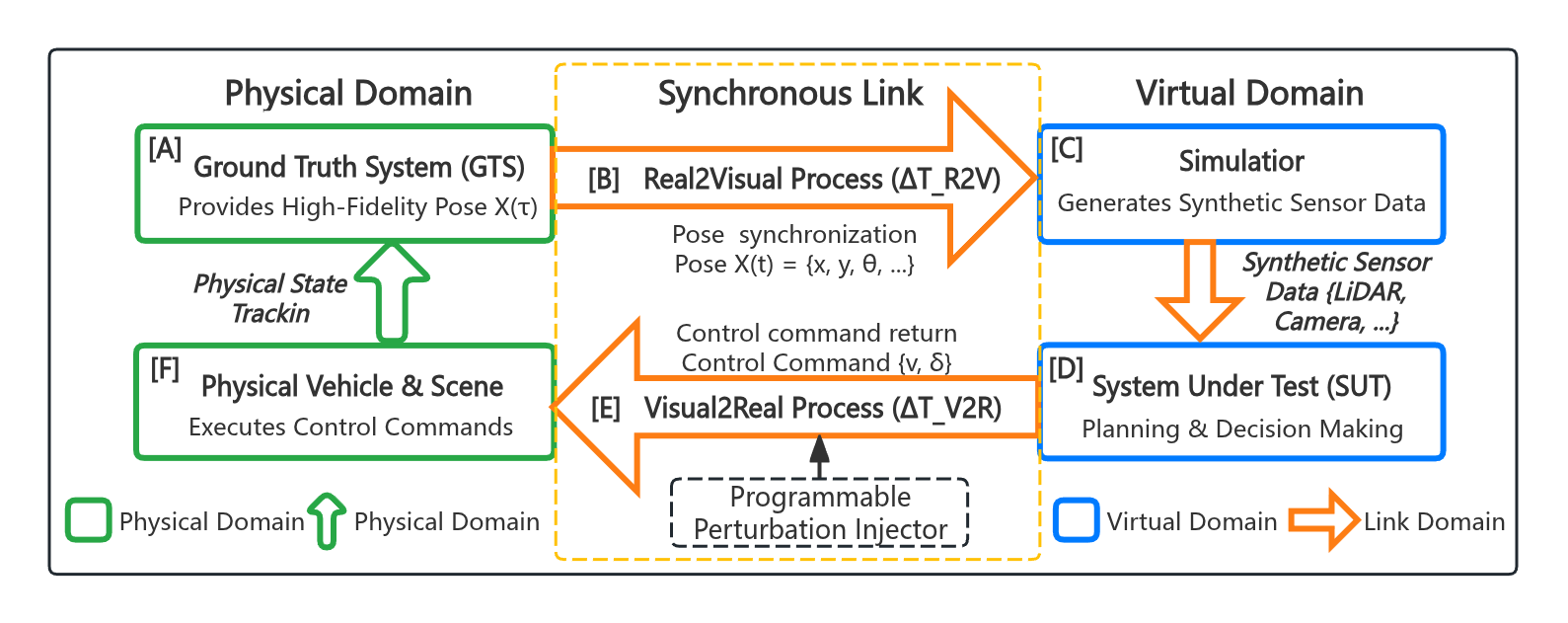}
    \caption{MMRHP System implementation and data pathways within HIL loop. }
    \label{fig:arch}
\end{figure*}

\subsection{MMRHP: Architecture and Implementation}
\label{sec:arch_and_impl}

To support the proposed measurement framework, we designed and implemented the MMRHP. Its core is a co-simulation architecture that bridges a real-world physical environment with its virtual counterpart, an overview of which is presented in Fig.~\ref{fig:panoramic}.

% ----------------- Figure: conceptual_arch -------------------
\begin{figure}[!h]
    \centering
    \includegraphics[width=0.8\columnwidth]{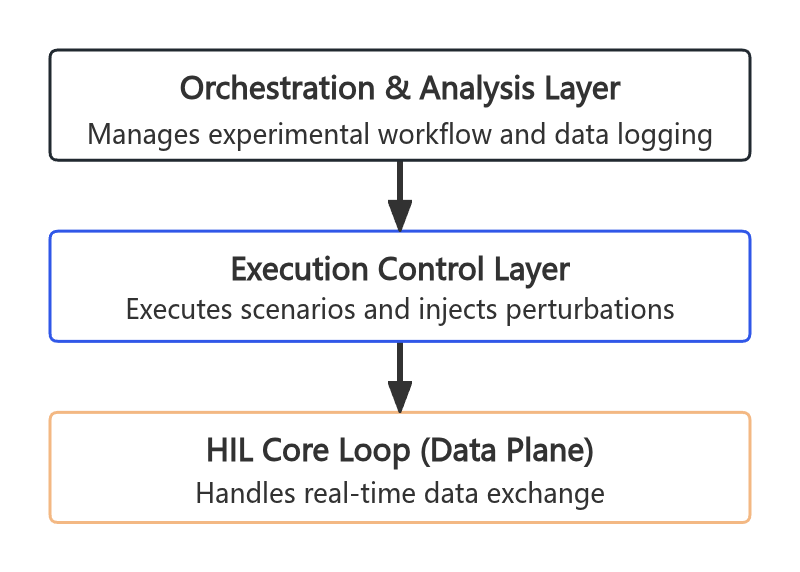} 
    \caption{The logical management and execution hierarchy of the MMRHP.}
    \label{fig:conceptual_arch}
\end{figure}

The management and execution of this architecture are structured logically as depicted in Fig.~\ref{fig:conceptual_arch}. An overarching \textbf{Orchestration \& Analysis Layer} governs the entire experimental lifecycle, enabling standardized, automated execution of the three-stage workflow. This layer delegates high-level commands to the \textbf{Execution Control Layer}, which is responsible for the deterministic, real-time execution of test scenarios and the injection of programmable perturbations. Both of these control layers operate upon the foundational \textbf{HIL Core Loop (Data Plane)},  which is physically constituted by three interconnected domains: the physical domain, the virtual domain, and the data links. The detailed implementation and data pathways within this loop are depicted in Fig.~\ref{fig:arch}.

\subsubsection{Physical Domain: Actuation and Ground Truth}
\label{sec:physical_domain}
The physical domain grounds the simulation in reality through tangible action and measurement. This is achieved via two primary components: a physical actuator and an external ground truth system (GTS). 

The actuator is a \textbf{1:28 scale Ackermann-steering vehicle [F]}, equipped with an onboard microcontroller that receives wireless commands via a 2.4 GHz transceiver. A key design element is a non-linear command mapping model implemented in software, which translates the SUT's idealized commands (e.g., velocity in m/s) into calibrated low-level PWM signals. This model is crucial for compensating for chassis dynamics and ensuring predictable physical execution, the fidelity of which is characterized in Sec.~\ref{sec:validation_actuator}.

Concurrently, the vehicle's real-world trajectory is tracked by a \textbf{high-precision external GTS [A]}, which provides an independent measurement baseline. Accurately implementing reference path-based measurement requires a method-independent and sufficiently accurate external ground truth system. In outdoor scenarios, this goal is often achieved through complex sensor fusion, using vehicle-mounted sensors such as RTK-GNSS\cite{vedder2018low}. For indoor platforms, motion capture equipment such as Vicon and OptiTrack is usually used. To balance accuracy with cost, we selected the low-cost Lighthouse 2.0 and Vive Tracker 3.0 solution. However, this consumer-grade system inherently suffers from signal noise, pose jitter, and insufficient accuracy to serve directly as a ground truth\cite{merker2023measurement,hsiao2022multi}. Therefore, we designed and implemented a complete data acquisition and processing framework specifically to overcome these limitations and produce a high-fidelity dataset.

The hardware for this framework comprises an HTC Vive Tracker 3.0 ($\approx 50$~Hz) rigidly connected to a calibration tool with dual orthogonal, high-precision single-point LiDARS ($\approx 10$~Hz). To fuse these asynchronous data streams, we use multi-threaded acquisition with a common monotonic timestamp. Subsequently, we use the arrival time $\tau_i$ of the low-frequency LiDAR as a reference to perform causal interpolation on the high-frequency Tracker's pose in $SE(3)$, generating temporally synchronized data pairs. These pairs are then refined through a pipeline of static, attitude-stability, and Savitzky-Golay filters. The resulting data train a hierarchical hybrid registration model $f$, of the form:

\begin{equation}
\label{eq:hybrid_model}
f(\mathbf{p}) = \mathbf{A}\mathbf{p} + \mathbf{t} + g(\mathbf{p}; \theta) 
\end{equation}
where $\mathbf{p}$ is the original input coordinate. This model first captures global linear distortions with an affine transformation (defined by matrix $\mathbf{A}$ and translation vector $\mathbf{t}$), and then uses a Multi-Layer Perceptron (MLP) $g(\mathbf{p}; \theta)$, defined by parameters $\theta$, to perform non-linear local correction of the residuals.

The detailed validation of this system's accuracy, demonstrating its ability to meet the required centimeter-level standard, is presented in Section~\ref{sec:valid_a_gts}.

\subsubsection{Virtual Domain: Simulation and Decision-Making}
\label{sec:virtual_domain}
Pose data from the physical domain enters the virtual domain, which comprises the simulator and the System Under Test (SUT). Our platform architecture relies on the \textbf{CARLA simulator [C]} operating in a \textbf{synchronous mode}. 
 
CARLA is responsible for rendering a configurable suite of virtual sensors based on the virtual vehicle's state. The \textbf{SUT [D]}, in our case Autoware, operates within a well-defined 'sandbox' environment. Its inputs are exclusively the ROS2 topics published by the platform (e.g., virtual sensor data), and its sole output is a standardized control command. This strict architectural contract enables true black-box measurement, ensuring that observed performance is attributable to the SUT's algorithms.

% --- Figure: Calibration Data Visualization ---
\begin{figure*}[!t]
\centering
\subfloat[]{\includegraphics[width=2.1in]{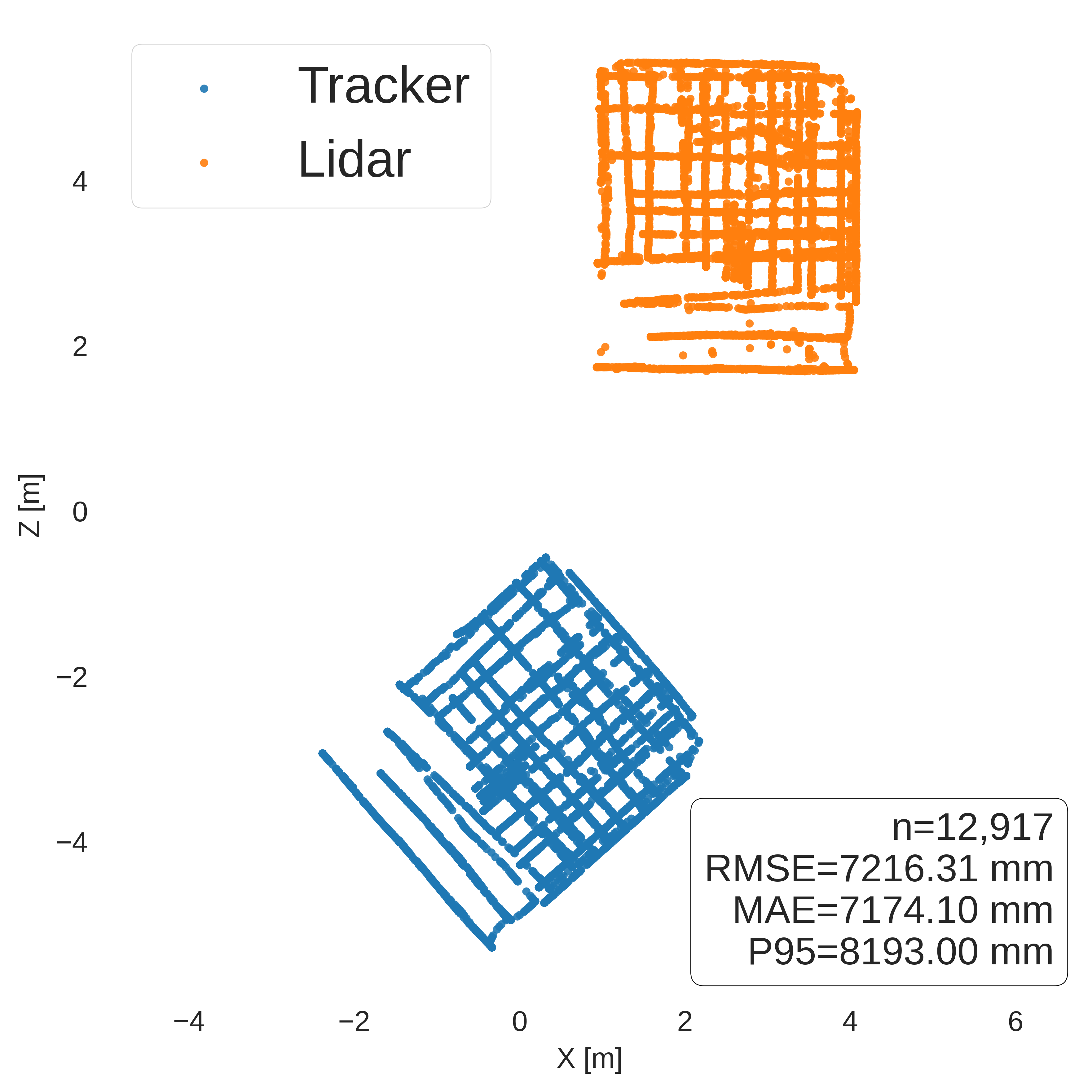}\label{fig_4_1a}} % 方括号留空
\hfil
\subfloat[]{\includegraphics[width=2.1in]{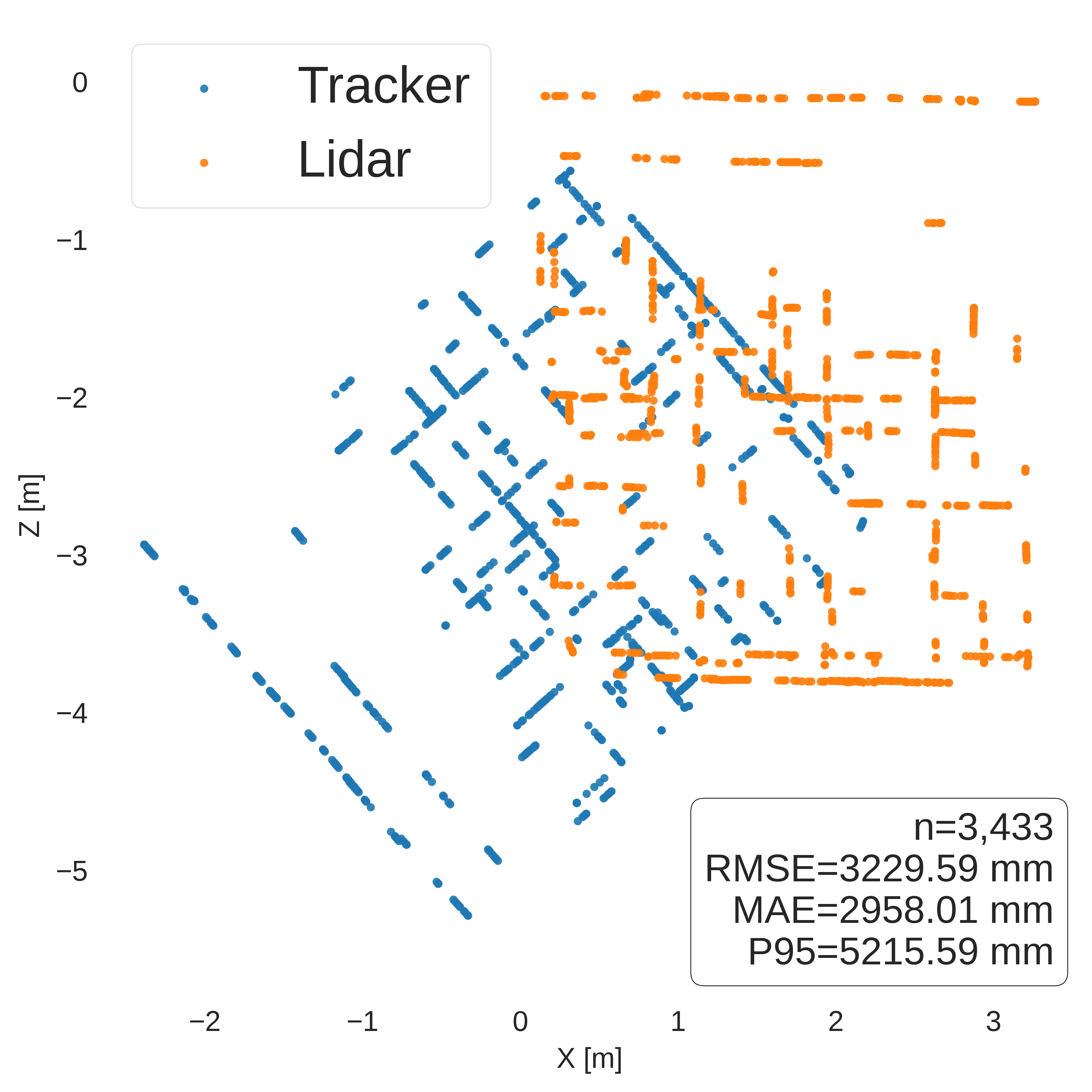}\label{fig_4_1b}} % 方括号留空
\hfil
\subfloat[]{\includegraphics[width=2.1in]{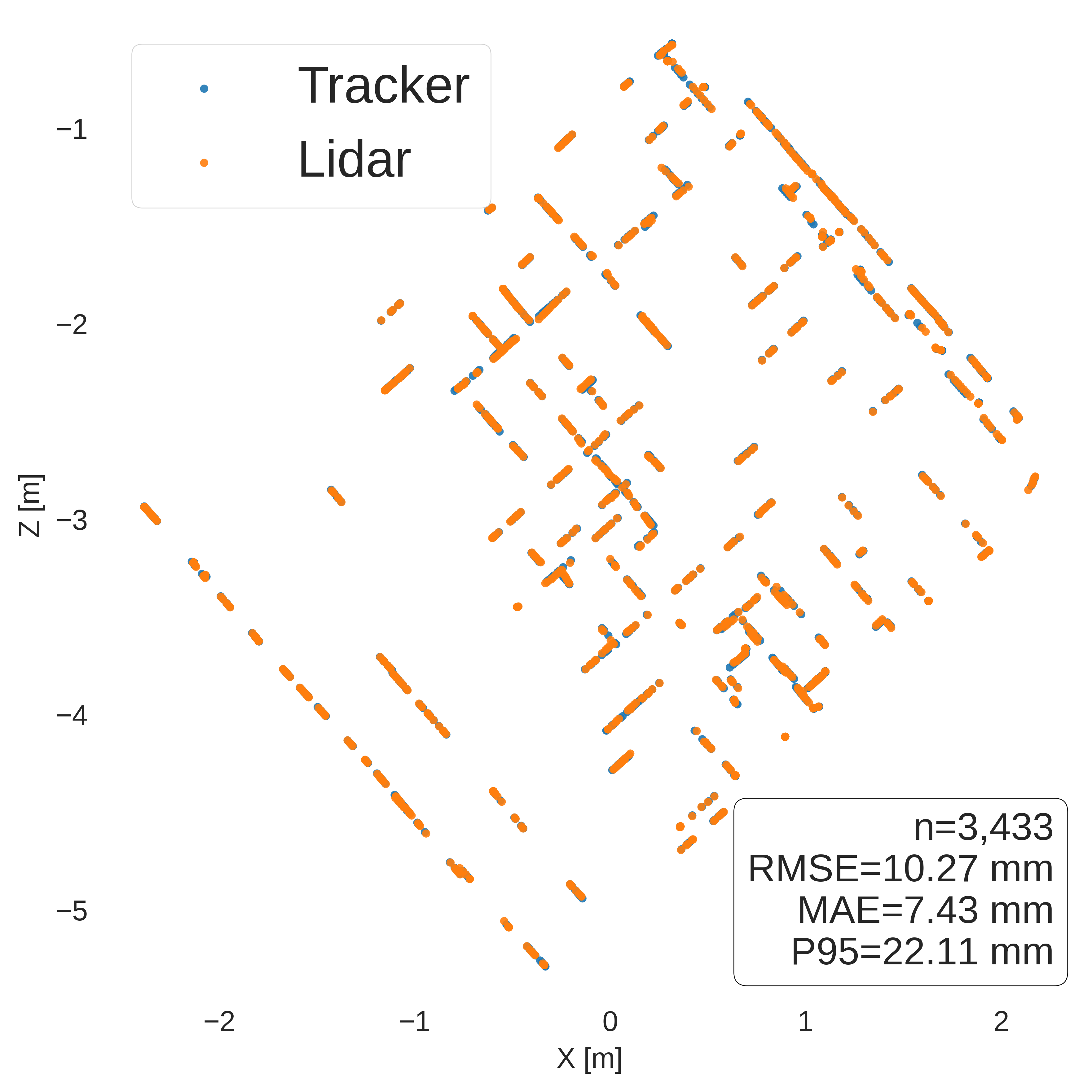}\label{fig_4_1c}} % 方括号留空
\caption{
Visualization of the calibration pipeline and alignment results. 
(\textbf{a})Raw data from the Tracker (blue) and LiDAR (orange) sensors, illustrating the large initial misalignment (RMSE $>7000$ mm).
(\textbf{b})The 3,433 matched point pairs after preprocessing and coarse alignment, still showing significant residual error (RMSE $\approx 3230$ mm). 
(\textbf{c})Final alignment of the test set data after correction by the proposed hybrid model, achieving a precise overlay and a final RMSE of 10.27 mm.}\label{fig:calib_data}
\end{figure*}

\subsubsection{HIL Core Loop: Data Links and Timing Control}
\label{sec:l_hil_core_loop}
Connecting the physical and virtual domains are two asymmetric data links. The core of this linkage is a strict timing model where all key events are timestamped in a unified monotonic clock domain $\tau$. This provides a reliable and verifiable data foundation for the latency decomposition analysis defined in Sec.~\ref{sec:latency_decomposition}, with the complete timing chain depicted in Fig.~\ref{fig:timing_diagram}.

The \textbf{Real-to-Virtual (R2V) link [B]} transfers pose data from the GTS [A] to the simulator [C]. It is implemented as a dedicated ROS2 node that calls the simulator's API to forcibly synchronize the virtual vehicle's state with its physical counterpart. To enable precise latency measurement, this node timestamps key events---namely, the ingestion of a new pose and the completion of the API call---using the host's monotonic clock. Each message is also tagged with a unique \textbf{correlation ID}, ensuring end-to-end traceability and providing the necessary data for the analysis defined in Sec.~\ref{sec:latency_decomposition}.

The \textbf{Virtual-to-Real (V2R) link [E]} transmits control commands from the SUT [D] to the physical actuator [F]. Its most critical feature is a \textbf{programmable perturbation injector}, implemented as a software-based scheduler. This module intercepts commands and, based on a configuration file, introduces configurable delays, jitter, or packet loss, which is critical for conducting the sensitivity analysis defined in our workflow (Sec.~\ref{sec:workflow}). After the perturbation stage, commands are transmitted through a two-tier communication system: they are first sent via a wired connection to a stationary communication gateway, which then wirelessly relays the final payload to the vehicle's onboard microcontroller. This architecture ensures reliable command transmission while decoupling the host PC from the mobile agent.

% --- Platform Validation Section ---
\section{Platform Validation: Establishing a Credible Measurement Cornerstone}
\label{sec:validation}
The core task of this section is to systematically validate the inherent reliability of the MMRHP as a measurement platform. This section decouples the MMRHP from the System Under Test (SUT) to quantitatively evaluate its foundational capabilities across three distinct domains:

\begin{enumerate}
    \item \textbf{Spatial Metrology (Sec. \ref{sec:valid_a_gts}):} We first establish the platform's ability to measure physical position by rigorously quantifying the accuracy and robustness of the external \textbf{Ground Truth System [A]}.
    \item \textbf{Temporal Metrology (Sec. \ref{sec:valid_b_e_link}):} Next, we characterize the end-to-end timing behavior of the entire HIL information loop. This analysis inherently quantifies the performance of all constituent components, including the data links \textbf{[B] \& [E]} and the virtual domain (\textbf{simulator [C]} and \textbf{SUT interface [D]}).
    \item \textbf{Physical Execution Fidelity (Sec. \ref{sec:validation_actuator}):} Finally, we assess the fidelity of the \textbf{physical vehicle [F]} by characterizing its dynamic response, allowing us to distinguish algorithmic errors from hardware limitations.
\end{enumerate}

\subsection{Spatial Metrology Validation: Ground Truth System [A]} \label{sec:valid_a_gts}
%------------------------------------------------------------------

This section validates the precision and robustness of the platform's external ground truth system. Through a series of experiments, we first confirm its centimeter-level accuracy, then assess its uncertainty and consistency under different conditions. The results verify that the system meets the requirements outlined in Sec. \ref{sec:arch_and_impl}.

\subsubsection{Core Accuracy Validation: Hybrid Model vs. Baseline Method}
To quantify the final accuracy of the system, we first conducted a core accuracy validation experiment. Using the calibration tool designed in \ref{sec:arch_and_impl}, we systematically collected over 12000 raw data points pair within a workspace of approximately 4.2m x 4.2m, with the spatial distribution shown in Fig.~\ref{fig:calib_data}(a). Subsequently, following the data processing pipeline detailed in Sec.~\ref{sec:physical_domain}, we pre-processed the raw data to obtain a total of 3433 high-quality coordinate matching pairs $\{p_T, p_L\}$(where $p_T$ and $p_L$ are the coordinates of the same point in the Tracker and Lidar coordinate systems, respectively), as shown in Fig.~\ref{fig:calib_data}(b), which will be used for subsequent model training and evaluation.

We randomly split the matched point pairs into an 80\% training (2746 points) and 20\% test (687 points) set to assess generalization. On the training set, we prepared our proposed hybrid model (global affine + MLP residual) and a standard SVD baseline. Upon evaluation with the  test data, our model demonstrated a better ability to correct non-linear distortions, achieving the near-perfect alignment with ground truth LiDAR data visualized in Fig.~\ref{fig:calib_data}(c).

The performance of the two methods on the independent test set is shown in Table \ref{tab:accuracy_comparison}. The table reports the Root Mean Square Error (RMSE), Mean Absolute Error (MAE), and the 95th percentile error (P95).

% --- Table Accuracy Comparison ---
\begin{table}[!h]
\caption{Coordinate Registration Method Accuracy Comparison (All metrics are in mm). \label{tab:accuracy_comparison}}
\centering
\renewcommand{\arraystretch}{1.2}
% Use tabularx and set its width to \columnwidth
\begin{tabularx}{\columnwidth}{l *{3}{>{\centering\arraybackslash}X}}
\toprule
\textbf{Metric} & \textbf{\makecell{RAW after \\ Filter}} & \textbf{SVD Baseline} & \textbf{Our Model} \\
\midrule
RMSE & 3229.59 & 16.85 & \textbf{10.27} \\
MAE  & 2958.01 & 14.78 & \textbf{6.87} \\
P95  & 5215.59 & 30.07 & \textbf{22.11} \\
\bottomrule
\end{tabularx}
\end{table}

As seen from the table, our proposed hybrid registration method achieves an RMSE of \textbf{10.27 mm} on the independent test set, with its accuracy improving by approximately 43.5\% compared to the SVD baseline method. This result strongly demonstrates the necessity and effectiveness of compensating for the non-linear distortions of consumer-grade localization hardware and clearly indicates that our external ground truth system has successfully met the preset centimeter-level accuracy target.

\subsubsection{Robustness Analysis: Uncertainty Quantification and Spatial Consistency}
\begin{figure}[!t]
\centering
\includegraphics[width=0.75\linewidth]{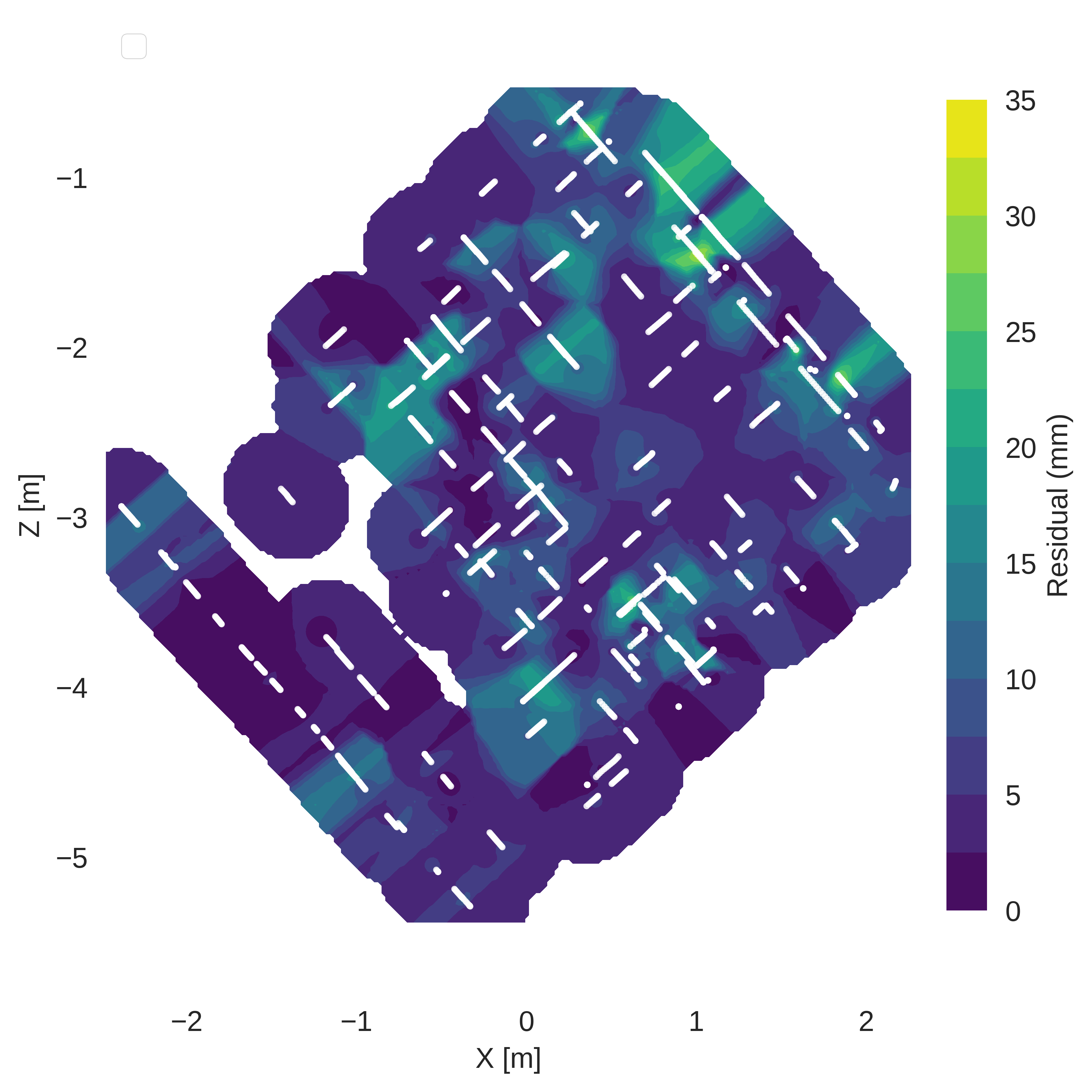}
\caption{Spatial distribution of coordinate registration residuals on the test set. Each point represents a sample, and its color indicates the Euclidean distance error after registration.}
\label{fig:residual_heatmap}
\end{figure}

To further investigate the model's robustness, we analyzed the spatial consistency of the registration residuals on the test set. Fig.~\ref{fig:residual_heatmap} visualizes the Euclidean distance error for each test point across the workspace. The heatmap clearly shows that the error remains low (approx. 10mm) and is uniformly distributed within the core working area. A slight increase in error appears only at the \textbf{far corners}, which is a known characteristic of the consumer-grade Vive system's base station coverage. Importantly, we \textbf{did not observe systematic uncaptured distortions in the primary workspace}. This result provides strong evidence for our hybrid model's effectiveness in correcting complex, non-linear spatial errors.

% --- Figure: Residual Heatmap ---

\subsection{Temporal Metrology Characterization of the HIL Loop}\label{sec:valid_b_e_link}
This section characterizes the timing properties of the entire HIL information loop, establishing a reliable temporal baseline. By analyzing the constituent latencies of the Real-to-Virtual (R2V) and Virtual-to-Real (V2R) links, we inherently quantify the performance of the underlying communication infrastructure \textbf{[B] \& [E]} and the virtual domain (\textbf{simulator [C]} and \textbf{interface [D]}).

All experiments were conducted without SUT intervention to measure only the platform's inherent latency. Timestamps were logged in a single monotonic clock domain ($\tau$). Based on $N=1860$ steady-state samples, our analysis reveals an asymmetric latency profile.

% --- TABLE: Latency Characteristics ---
\begin{table}[!h]
\caption{MMRHP Core Link Latency Characteristics}
\label{tab:latency_char}
\centering
\renewcommand{\arraystretch}{1.2}
\begin{tabularx}{\columnwidth}{@{} X S[table-format=2.2]
                                 S[table-format=2.2]
                                 S[table-format=1.2] @{}}
\toprule
\textbf{Latency Component} & {\textbf{Mean (ms)}} & {\textbf{Std Dev (ms)}} & {\textbf{CV}} \\
\midrule
\textbf{$\Delta T_{\text{R2V}}$ (End-to-End)} & \bfseries 36.58 & \bfseries 23.46 & \bfseries 0.64 \\
\quad\textit{-- $\Delta T_{\text{ingest}}$} & 0.26 & 0.11 & 0.42 \\
\quad\textit{-- $\Delta T_{\text{adv}}$} & 28.68 & 23.22 & 0.81 \\
\quad\textit{-- $\Delta T_{\text{sense}}$} & 7.64 & 2.66 & 0.35 \\
\midrule
\textbf{$\Delta T_{\text{V2R}}$ (End-to-End)} & \bfseries 8.58 & \bfseries 1.2 & \bfseries 0.14 \\
\bottomrule
\end{tabularx}
\end{table}

\subsubsection{V2R Link: High Stability}
\begin{figure}[!t]
    \centering
    
    \subfloat[V2R Latency over Time]{
        \includegraphics[width=0.95\linewidth]{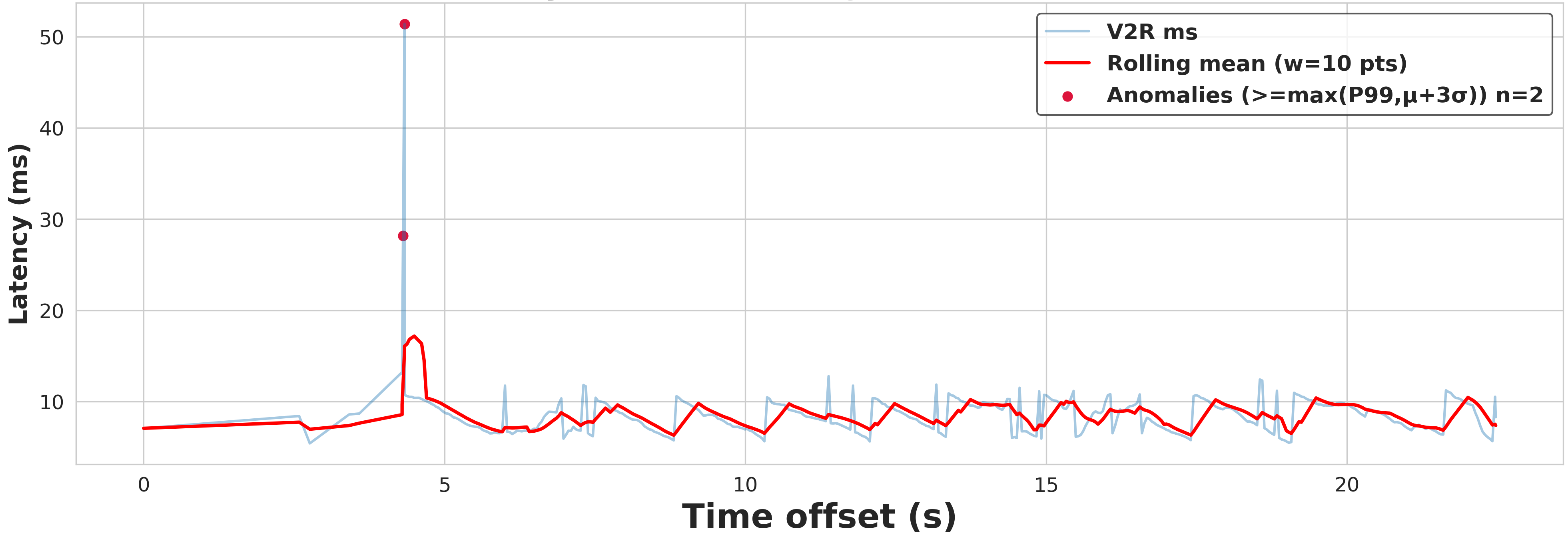}
        \label{fig:v2r_ts_latency}
    }
    \vspace{2mm} 

    \subfloat[V2R Jitter over Time]{
        \includegraphics[width=0.95\linewidth]{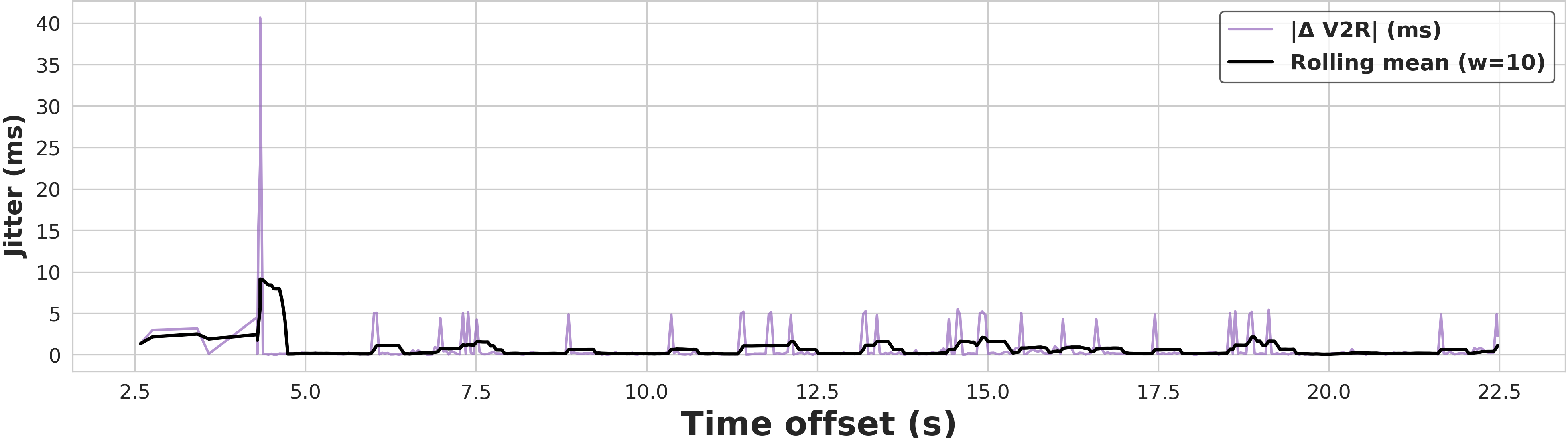}
        \label{fig:v2r_ts_jitter}
    }
    \vspace{2mm} 

    \subfloat[V2R Latency Distribution]{
        \includegraphics[width=0.75\linewidth]{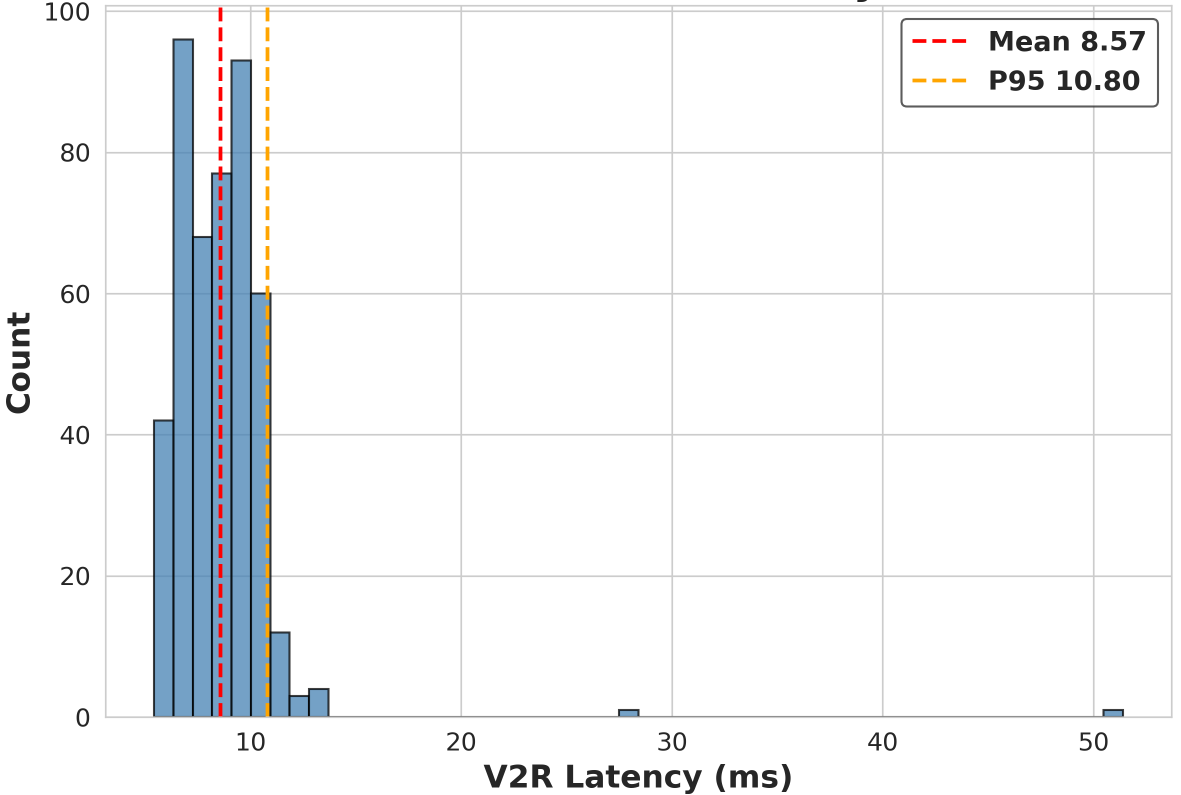}
        \label{fig:v2r_dist}
    }

    \caption{Temporal characteristics of the Virtual-to-Real (V2R) link. 
    }
    \label{fig:v2r_analysis_combined} 
\end{figure}

The V2R link, responsible for transmitting control commands, demonstrates exceptional stability. As visually confirmed in Fig.~\ref{fig:v2r_analysis_combined}, the latency exhibits a stable time series (Fig.~\ref{fig:v2r_ts_latency}), minimal corresponding jitter (Fig.~\ref{fig:v2r_ts_jitter}), and a tight, narrow probability distribution (Fig.~\ref{fig:v2r_dist}). This high predictability is quantitatively substantiated by the data in Table~\ref{tab:latency_char}, which shows a mean latency of \textbf{8.58 ms} and a low coefficient of variation (CV) of 0.14.

\subsubsection{R2V Link: Characterized Variability}
In contrast, the R2V link, which synchronizes the physical world to the virtual, shows significant variability. While its auditability remains intact, the total R2V latency has a high standard deviation and a CV of 0.64, as detailed in Table~\ref{tab:latency_char}.

\begin{figure}[t!]
    \centering
    
    \subfloat[Total R2V latency over time\label{fig:r2v_ts_full}]{%
        \includegraphics[width=\linewidth]{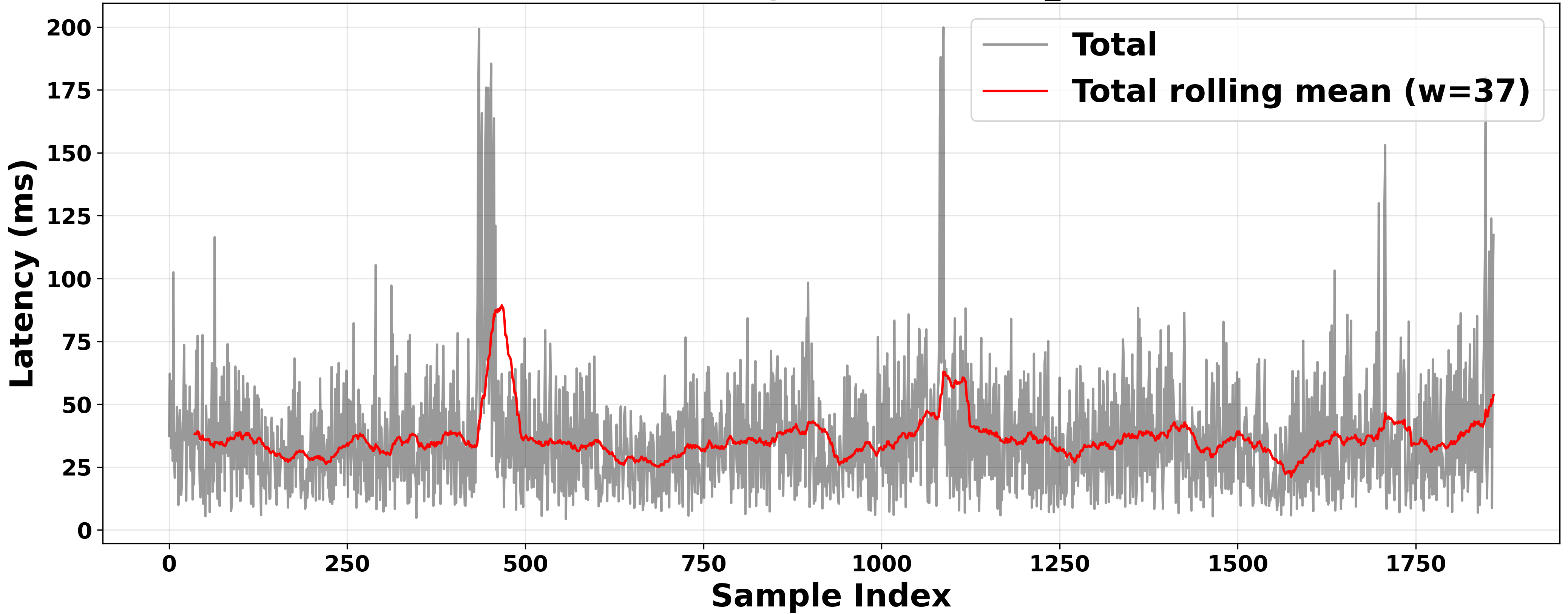}%
    }\\ 

    \subfloat[Total ($L_{\text{R2V}}$)\label{fig:r2v_dist_total}]{%
        \includegraphics[width=0.49\linewidth]{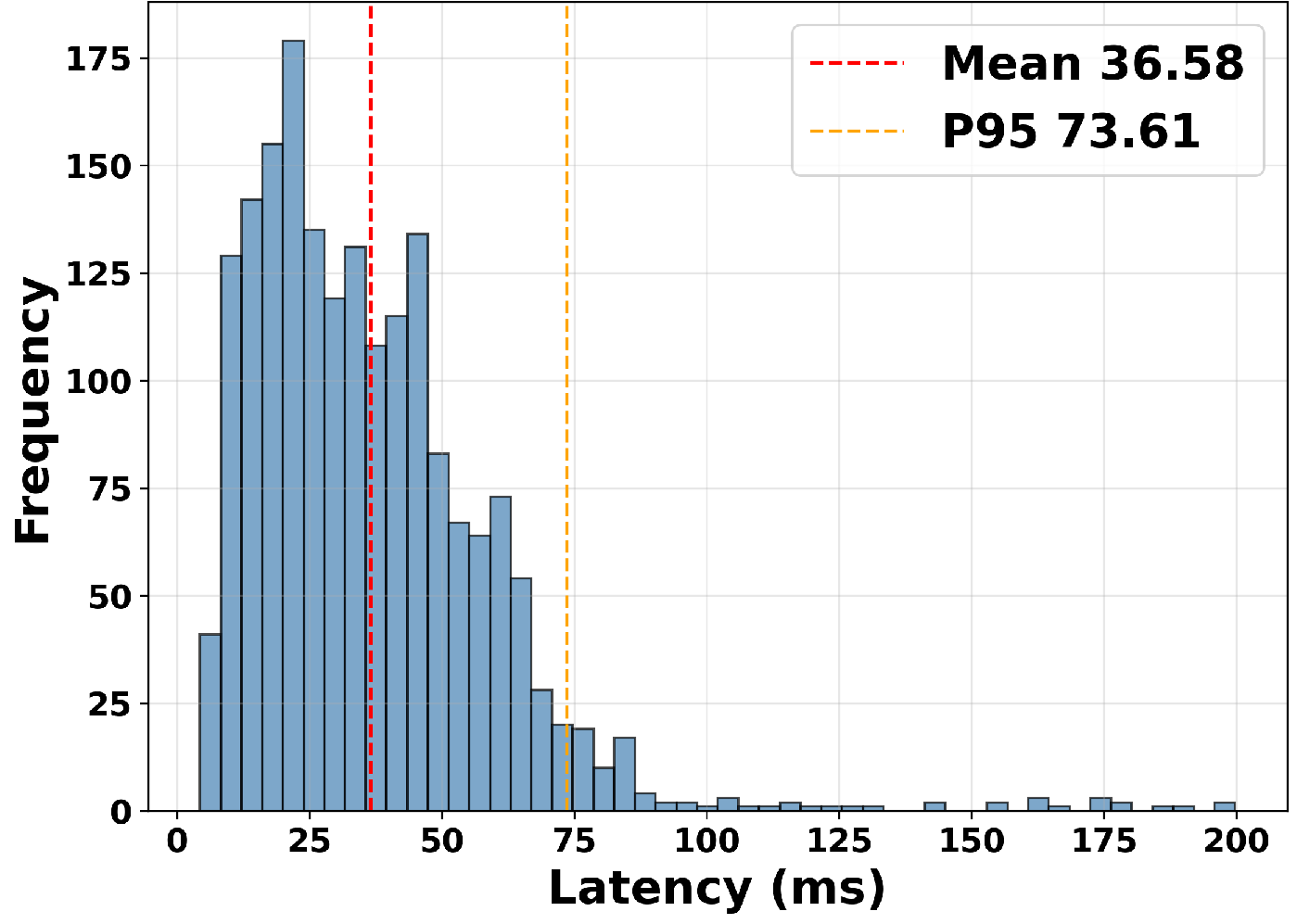}%
    }\hfill
    \subfloat[Advancement ($L_{\text{adv}}$)\label{fig:r2v_dist_adv}]{%
        \includegraphics[width=0.49\linewidth]{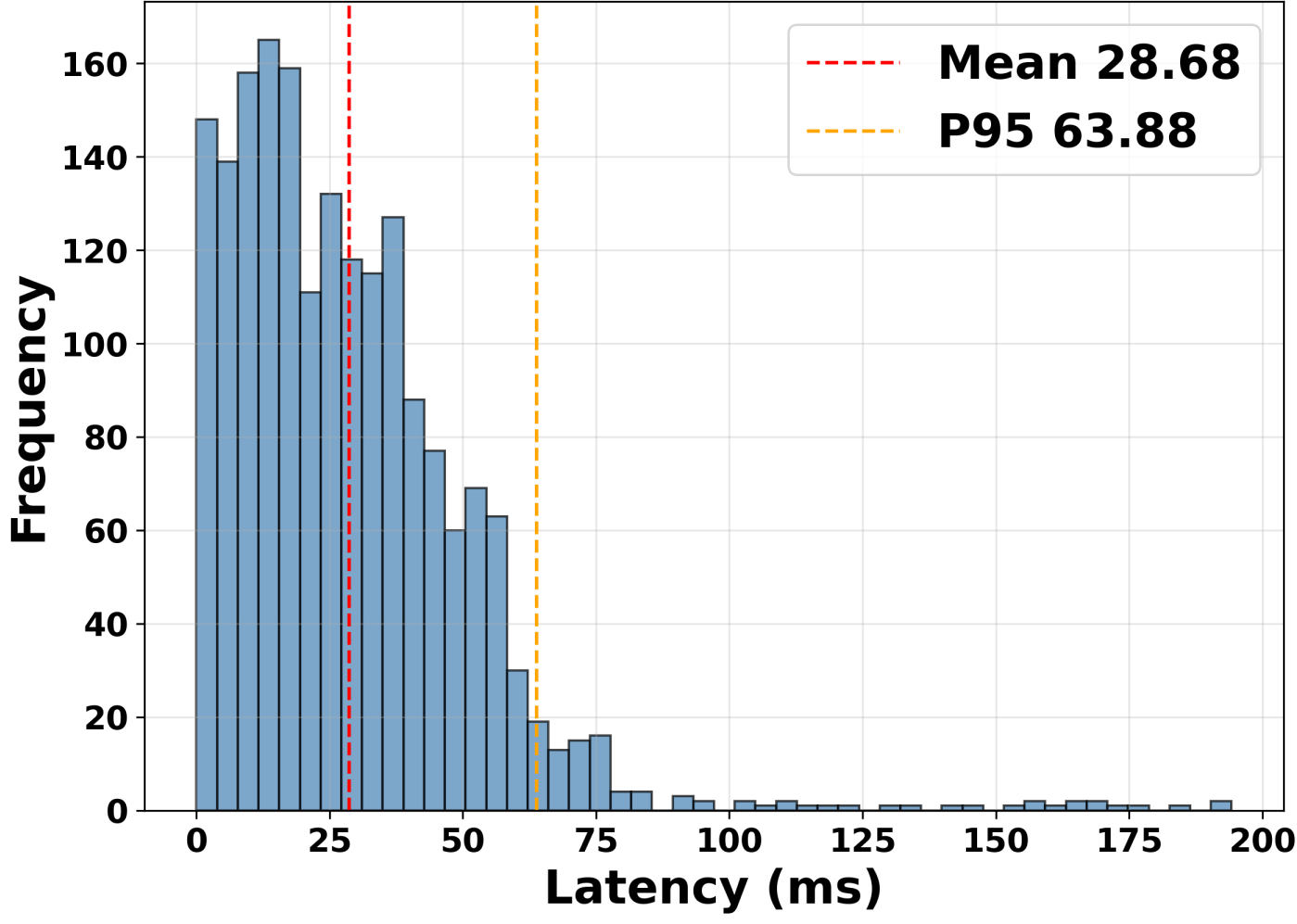}%
    }\\ 

    \subfloat[Ingest ($L_{\text{ingest}}$)\label{fig:r2v_dist_ingest}]{%
        \includegraphics[width=0.49\linewidth]{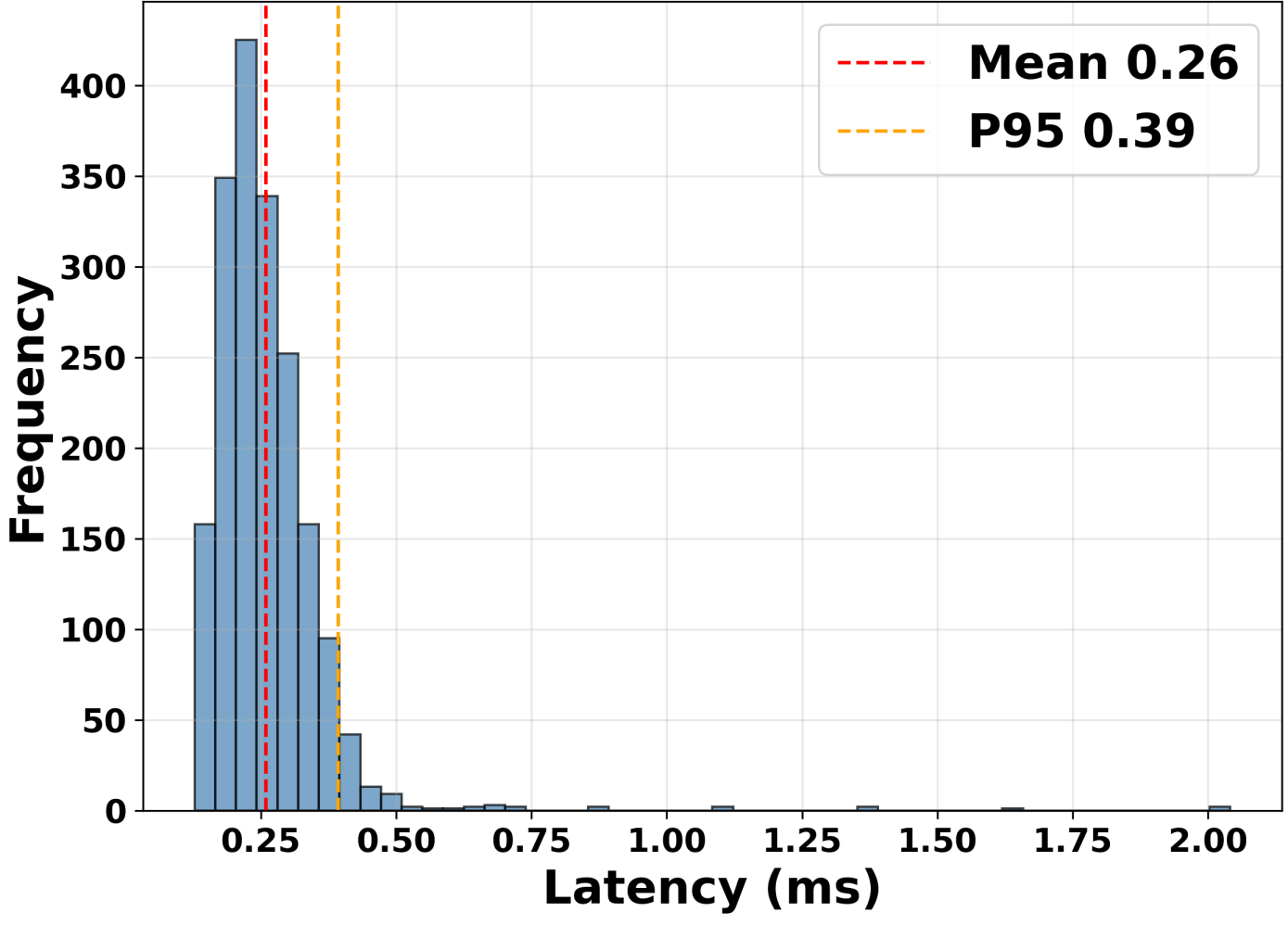}%
    }\hfill
    \subfloat[Sense ($L_{\text{sense}}$)\label{fig:r2v_dist_sense}]{%
        \includegraphics[width=0.49\linewidth]{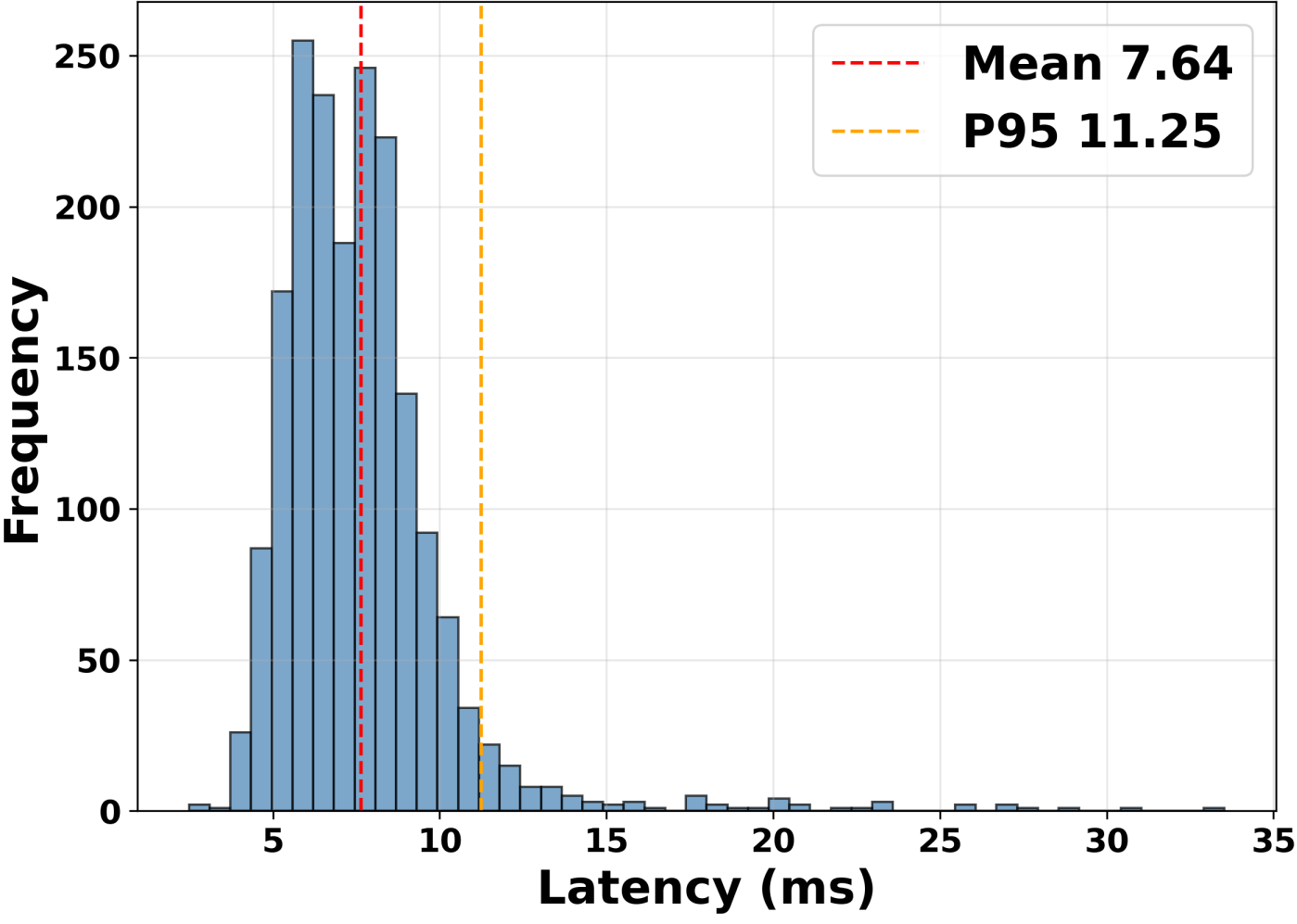}%
    }
    
    \caption{Latency analysis of the R2V link. (a) Total latency over time. (b-e) Latency distributions for the total link and its constituent stages.}
    \label{fig:r2v_analysis}
\end{figure}

Our latency decomposition model, visualized in Fig.~\ref{fig:r2v_analysis}, pinpoints the source of this instability. The long-tailed distribution of the \textbf{advancement stage ($\Delta T_{\text{adv}}$)} reveals that the CARLA \textbf{simulator [C]} is the primary contributor to latency jitter under our experimental load. This is a critical finding: rather than assuming simulator stability, our platform quantifies its real-world performance as a measurable component of the overall latency budget. The other stages, \emph{ingest} and \emph{sense}, remain highly stable. 

% =====  FIGURE 7 - v2r_latency 

% ================  R2V  =================

\subsection{Physical Execution Fidelity Characterization [F]}
\label{sec:validation_actuator}

To quantitatively establish the dynamic response baseline of the physical vehicle, a series of open-loop step command tests were conducted. This process is critical for decoupling the performance of the software-under-test (SUT) from the inherent limitations of the physical plant. The collected command-response data was fitted to a First-Order Plus Dead Time (FOPDT) model, whose high fidelity ($R^2 > 0.97$) confirms its suitability. The key parameters are summarized in Table~\ref{tab:fopdt_summary}.

The analysis reveals two distinct and critical characteristics of the testbed's dynamics.

% --- table of F ---
\begin{table}[!h]
\caption{Summary of vehicle dynamic response metrics}
\label{tab:fopdt_summary}
\centering
\footnotesize
\setlength{\tabcolsep}{5pt}
\renewcommand{\arraystretch}{1.15}
\begin{tabularx}{\columnwidth}{@{}l
>{\centering\arraybackslash}X
>{\centering\arraybackslash}X@{}}
\toprule
\textbf{Metric} & \textbf{Steering (lateral)} & \textbf{Velocity (longitudinal)} \\
\midrule
\multicolumn{3}{@{}l}{\textit{Temporal dynamics}} \\
System dead time ($L$, mean) & \textbf{7.2 ms} & 23.6 ms \\
System dead time ($L$, p95) & 35.2 ms & 92.3 ms \\
Time to 90\% response ($t_{90}$) & \textbf{0.85 s} & 0.53 s \\
\midrule
\multicolumn{3}{@{}l}{\textit{Steady-state fidelity}} \\
Process gain ($K$, median) & 0.26 & \textbf{2.11} \\
Model fit ($R^2$, median) & 0.990 & 0.978 \\
\bottomrule
\end{tabularx}
\end{table}

First, the steering channel analysis highlights a crucial distinction between communication latency and the vehicle's physical agility. The mean System Dead Time ($L$) is a mere \textbf{7.2 ms}, confirming a highly efficient control signal transmission. In stark contrast, the Time-to-90\%-Response ($t_{90}$) is \textbf{0.85 s}.This value is not merely actuator lag; rather, it quantifies the vehicle's \textbf{holistic cornering response} (the time from command to the stabilization of the vehicle's yaw angle). This 0.85 s measurement thus establishes a fundamental performance boundary for any control algorithm on this platform.

Second, the longitudinal channel, while dynamically faster ($t_{90}=0.53$ s), uncovers a significant calibration flaw. The median Process Gain ($K$) of \textbf{2.11} directly diagnoses a systemic over-response, where the vehicle's speed change is more than double the commanded value. This finding is not a failure, but a demonstration of the platform's diagnostic power, providing a clear, actionable path for correction by scaling the control mapping.

In essence, this characterization process ``calibrates'' our physical testbed. By quantifying core dynamic parameters such as the \textbf{0.85 s cornering response limit} and the \textbf{2.11 longitudinal gain error}, we establish a baseline. This baseline is indispensable for the unambiguous attribution of system-level errors during all subsequent SUT evaluations.

\subsection{Summary of Validated Platform Baseline}
\label{sec:validation_summary}
In conclusion, this chapter has systematically characterized the MMRHP as a measurement instrument. The platform provides a baseline with the following quantified properties:

\begin{itemize}
    \item \textbf{Spatial Metrology:} The ground truth system [A] provides a high-precision spatial reference with a validated accuracy of \textbf{10.27 mm} RMSE.
    \item \textbf{Temporal Metrology:} The HIL loop has a well-characterized, asymmetric latency profile. It features a highly stable V2R link (8.58 ms mean) and a variable but fully audited R2V link (36.58 ms mean), with the simulator [C] identified as the primary source of jitter.
    \item \textbf{Physical Fidelity:} The vehicle's dynamic response [F] is characterized, establishing a fundamental \textbf{cornering response limit of 0.85 s} ($t_{90}$) and uncovering a critical longitudinal \textbf{gain error of 2.11}, which demonstrates the platform's diagnostic power.
\end{itemize}

This comprehensive, component-wise validation establishes the MMRHP as a credible and trustworthy measurement instrument, providing a solid and fully characterized foundation for the SUT performance evaluation presented in the next chapter.

%fig：autoware-basline
\begin{figure*}[!t]
\centering
\subfloat[Trajectory Comparison]{\includegraphics[width=2.2in]{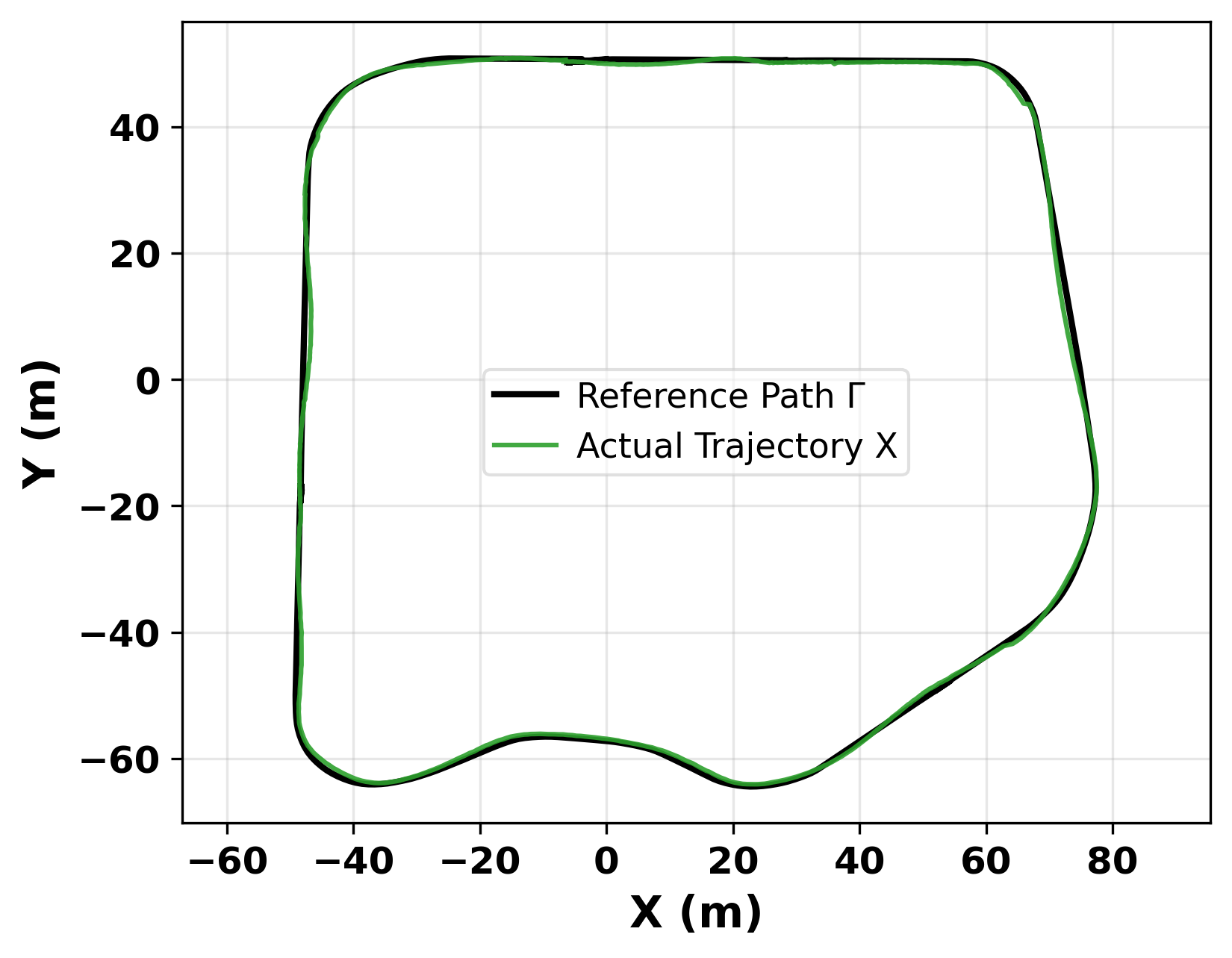}\label{fig:autoware_baseline_a}}
\hfil
\subfloat[Cross-Track Error Heatmap]{\includegraphics[width=2.2in]{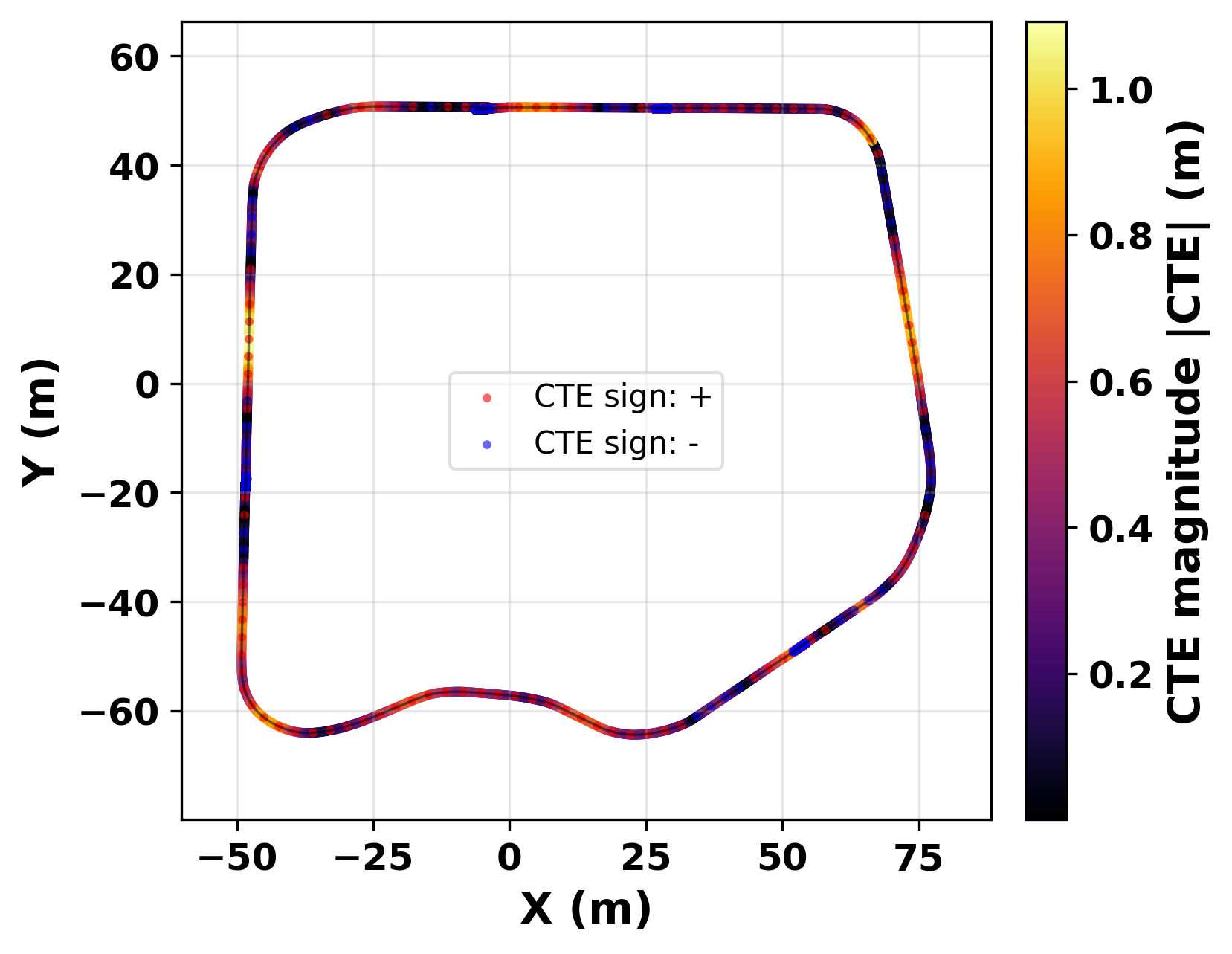}\label{fig:autoware_baseline_b}}
\hfil
\subfloat[Processing Latency Histogram]{\includegraphics[width=2.2in]{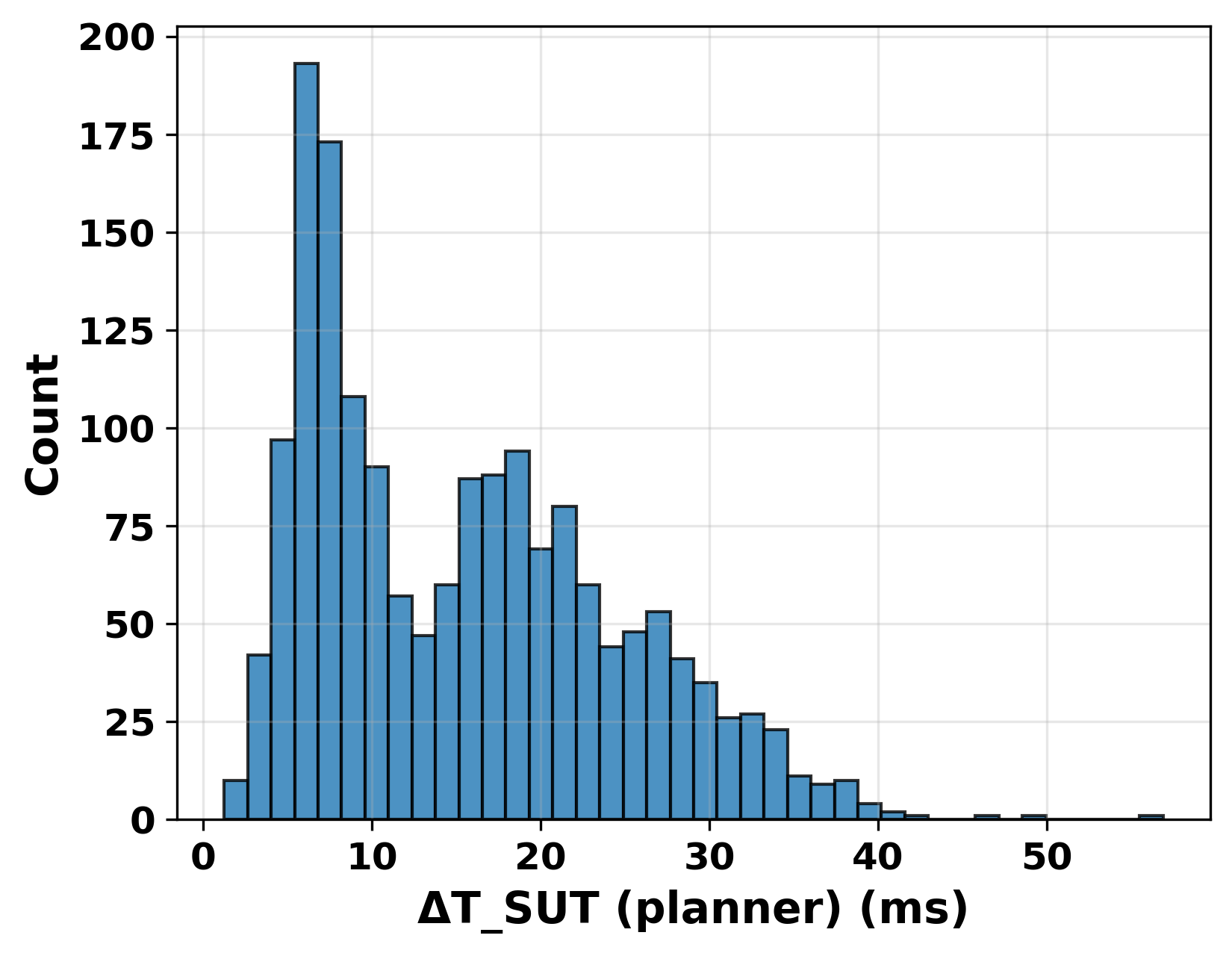}\label{fig:autoware_baseline_c}}
\caption{Autoware's baseline performance for a counter-clockwise path tracking task. (a) Comparison of the physical entity trajectory $X$ (green) with the reference path $\Gamma$ (black). (b) Distribution of Cross-Track Error (CTE) along the path, with redder colors indicating larger errors. (c) Histogram of Autoware's processing latency $\Delta T_{SUT}$.}
\label{fig:autoware_baseline}
\end{figure*}

\section{Case Study: A Three-Stage Diagnostic Methodology for Autoware Performance Evaluation}
\label{sec:case_study}
Having systematically validated the MMRHP as a reliable measurement instrument in the previous chapter, this chapter aims to demonstrate its practical value in analyzing a real, complex autonomous driving system. To this end, we use Autoware, a widely adopted open-source software stack, as the System Under Test (SUT) and apply the three-stage diagnostic methodology defined in Section \ref{sec:workflow}. This case study aims to specifically illustrate how the platform translates the various technical capabilities defined in Chapter \ref{sec:system_design} into interpretable insights into the SUT's performance.

\subsection{Stage 1: Quantitative Baseline Characterization of Autoware}

According to the methodology defined in Section \ref{sec:workflow}, the initial stage of the diagnostic process is to establish a performance baseline for the SUT. In this stage, we place Autoware in an idealized tracking task. The intervention of MMRHP is manifested in two core aspects: first, using its high-precision external ground truth system (Sec. \ref{sec:validation}), to accurately record the actual trajectory $X$ produced by the SUT-controlled physical vehicle; second, based on the spatial metric model defined in Section \ref{sec:measurement_framework}, to calculate the Cross-Track Error (CTE) of this trajectory relative to the preset reference path $\Gamma$.

%======================table Autoware Baseline Performance Metrics=================
\begin{table}[!h]
\caption{Autoware Baseline Performance Metrics}
\label{tab:5_1}
\centering
\renewcommand{\arraystretch}{1.2}
% Using tabularx for full width and S columns for decimal alignment.
\begin{tabularx}{\columnwidth}{@{} X S[table-format=2.4] 
                                 S[table-format=1.4] 
                                 S[table-format=2.4] @{}}
\toprule
\textbf{Performance Metric(abs)} & {\textbf{Mean}} & {\textbf{Std Dev}} & {\textbf{95th Pctl}} \\
\midrule
\textbf{CTE (m)} & 0.2539 & 0.2107 & 0.2262 \\
\textbf{$\Delta T_{\text{SUT}}$ (ms)} & 15.48 & 8.997 & 32.339 \\
\bottomrule
\end{tabularx}
\end{table}

The results of this process, summarized in Fig.~\ref{fig:autoware_baseline} and Table~\ref{tab:5_1}, establish a stable performance baseline for Autoware within our test architecture, characterized by a CTE RMSE of \textbf{0.2539m}, and a processing latency \textbf{$\Delta_{SUT}=15.48ms $}. However, beyond this aggregate measure, a deeper analysis of the data reveals two critical behavioral signatures.

In the spatial domain, the analysis reveals a systematic pattern (Fig.~\ref{fig:autoware_baseline}(b)): when navigating 90-degree turns, the vehicle exhibits a consistent \textbf{spatial undershoot} (corner-cutting). This behavior is characteristic of a system with significant response latency; the delayed turn-in forces an aggressive corrective steering action from the controller, causing the vehicle to cut sharply inside the corner's apex. Concurrently, in the temporal domain, the captured processing latency ($\Delta T_{\text{SUT}}$) exhibits a distinct \textbf{bimodal distribution} (Fig.~\ref{fig:autoware_baseline_c}). This non-Gaussian structure suggests the latency is not random noise but is instead driven by the planner operating under distinct computational loads.

\subsection{Stage 2: Analysis of Autoware's Robustness Boundary to System Latency}

%===========================fig：CTE RMSE vs. Injected Latency=======================
\begin{figure*}[!t]  
\centering
\subfloat[CTE RMSE vs. Injected Latency]{
    \includegraphics[width=0.49\textwidth]{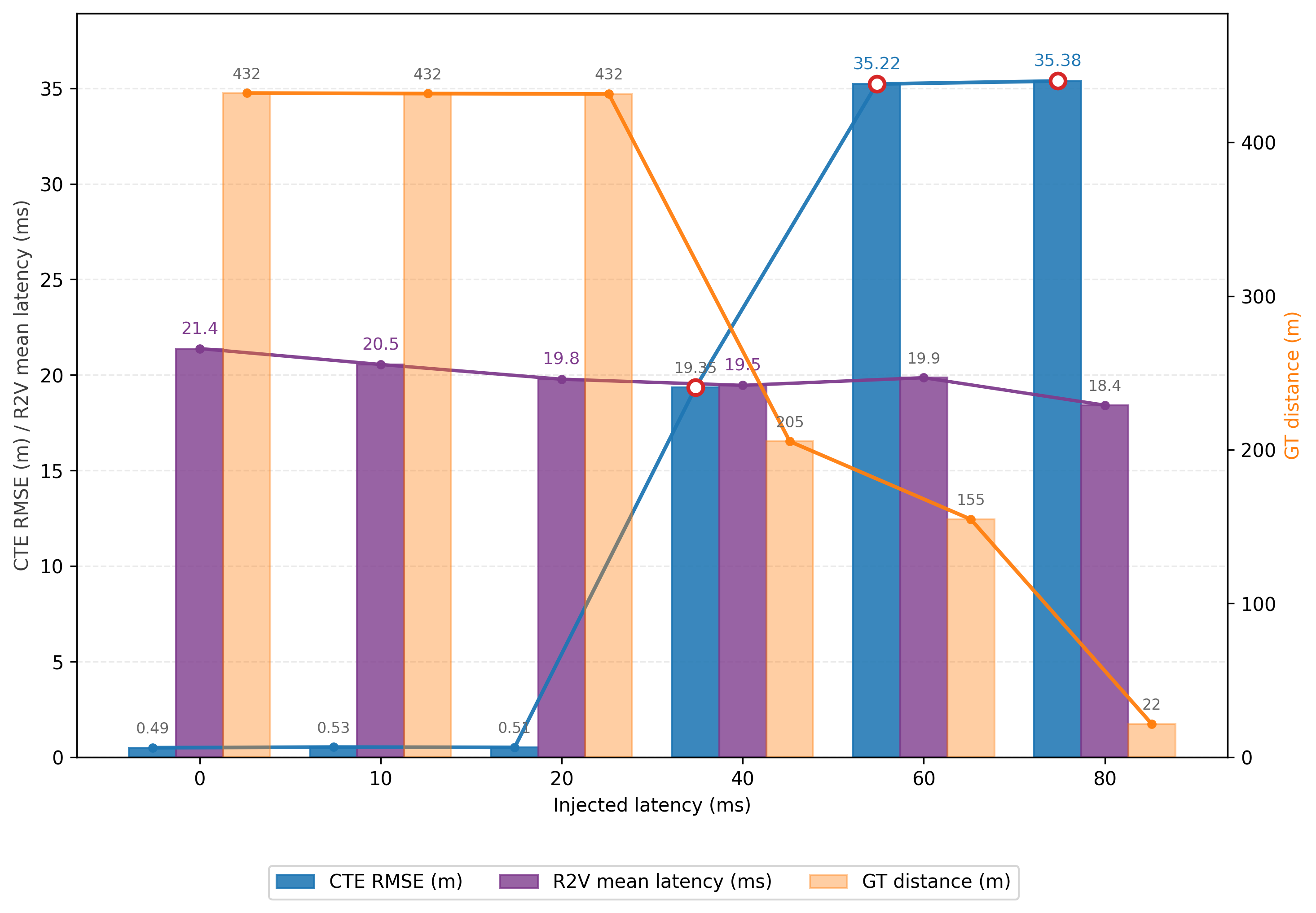}
    \label{fig:5_2a}
}\hspace{0.8cm}
\subfloat[Trajectories under Different Latencies]{
    \includegraphics[width=0.42\textwidth]{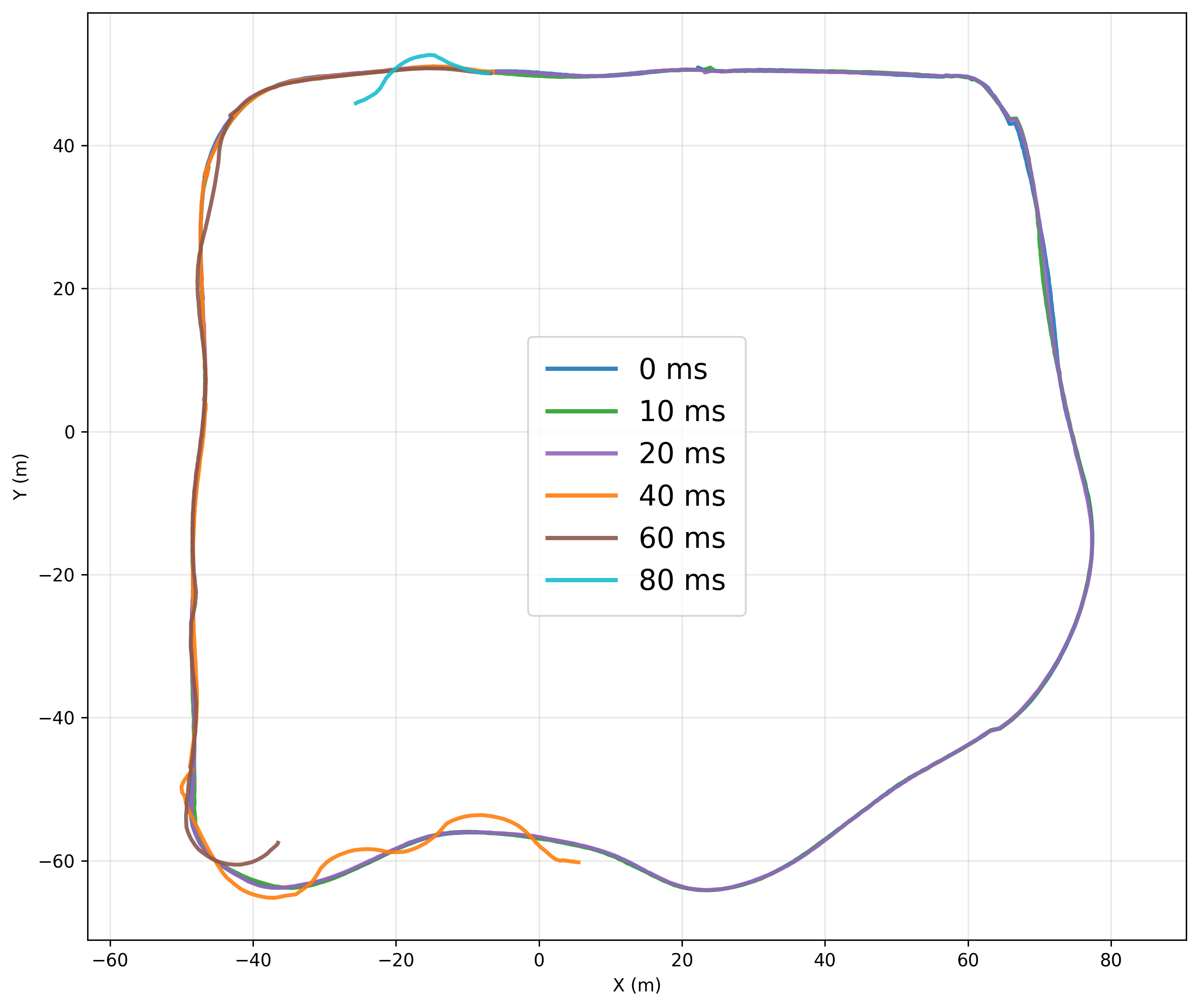}
    \label{fig:5_2b}
}

\caption{Analysis of Autoware's robustness boundary to latency. 
\textbf{(a)} Tracking error (CTE RMSE) and total distance traveled plotted against injected latency. 
\textbf{(b)} The corresponding vehicle trajectories, color-coded by the injected latency level.}
\label{fig:5_2}
\end{figure*}

After completing the baseline characterization, we proceed to the sensitivity analysis stage as defined in Section \ref{sec:workflow}. This stage aims to quantitatively determine how the SUT's performance degrades with specific, controlled perturbations and to identify its safe operational boundaries.

To achieve this, we utilize the programmable perturbation injector located in the V2R link, as described in Section \ref{sec:arch_and_impl}. While keeping the path tracking task unchanged, we programmatically inject additional delays in steps including `{0, 10, 20, 40, 60, 80}` ms on top of the system's inherent platform latency to probe the system's response.

The experimental results, shown in Fig.~\ref{fig:5_2}, reveal a striking non-linear response, which is more insightful than a simple linear degradation. Over the range of injected delays from 0 ms to 20 ms, Autoware's tracking performance, measured by CTE RMSE, exhibits \textbf{remarkable robustness}, with the error metric remaining stable at a low level of approximately 0.5 m. This indicates that the system is insensitive to low-to-moderate levels of latency jitter.

However, a sharp \textbf{performance cliff} is clearly observed when the injected latency increases from 20 ms to 40 ms. As shown in Fig.~\ref{fig:5_2a}, at an injected delay of 40 ms, the CTE RMSE catastrophically surges from 0.51 m to 19.35 m. This collapse in performance is visually corroborated by the trajectory plot (Fig.~\ref{fig:5_2b}): at 40 ms and higher latencies, the vehicle completely fails to negotiate the bottom-left turn, leading to a mission failure. This suggests that the system's control loop stability is entirely compromised at this latency level.

\subsection{Stage 3: Analysis of Autoware's Robustness to External Complexity}

\begin{figure*}[!t]
    \centering
    \subfloat[Rendered experimental environment.]{\includegraphics[width=0.48\textwidth]{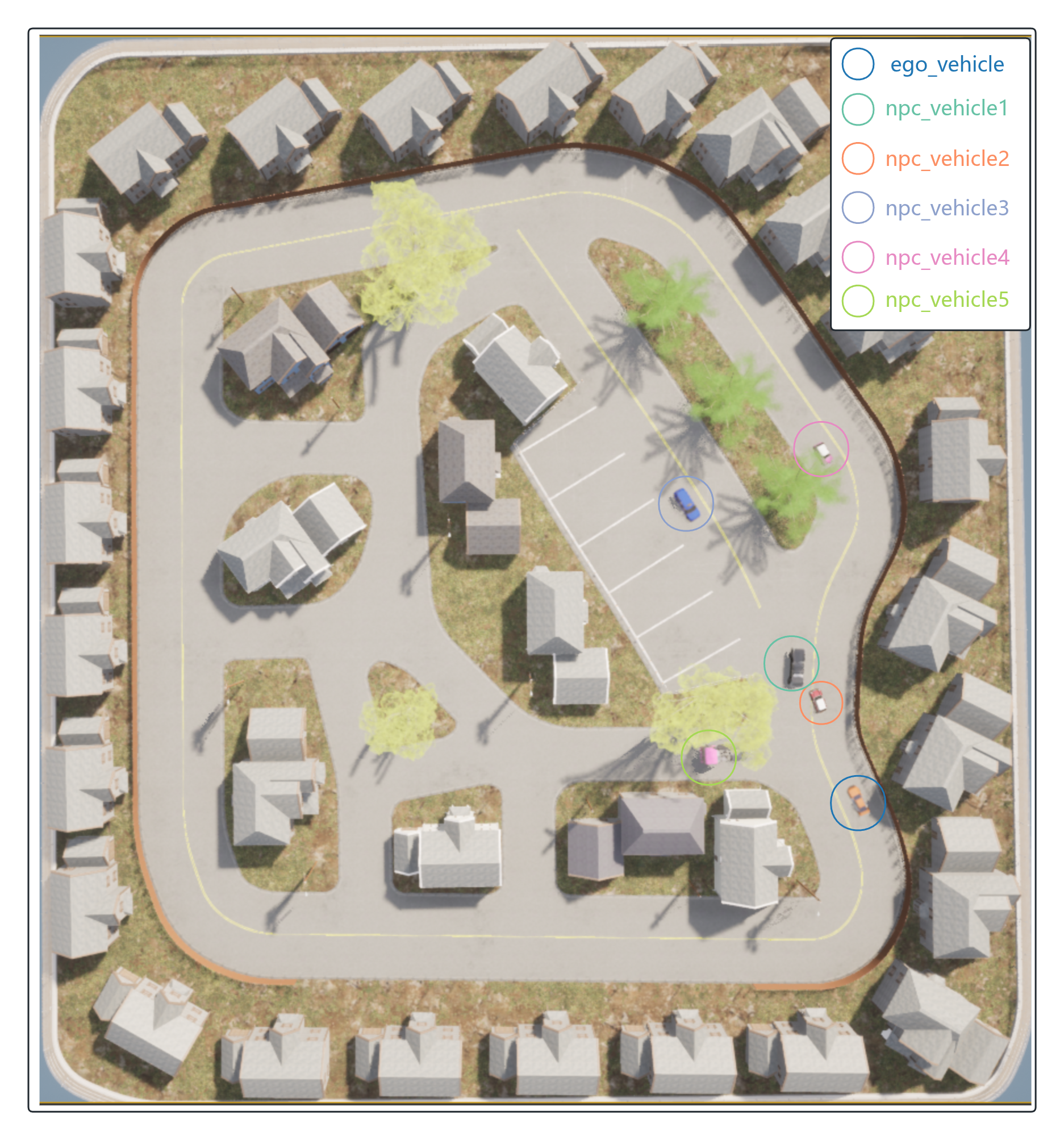}\label{fig:context_a}}
    \hfill 
    \subfloat[Global trajectories of the Ego and NPC vehicles.]{\includegraphics[width=0.48\textwidth]{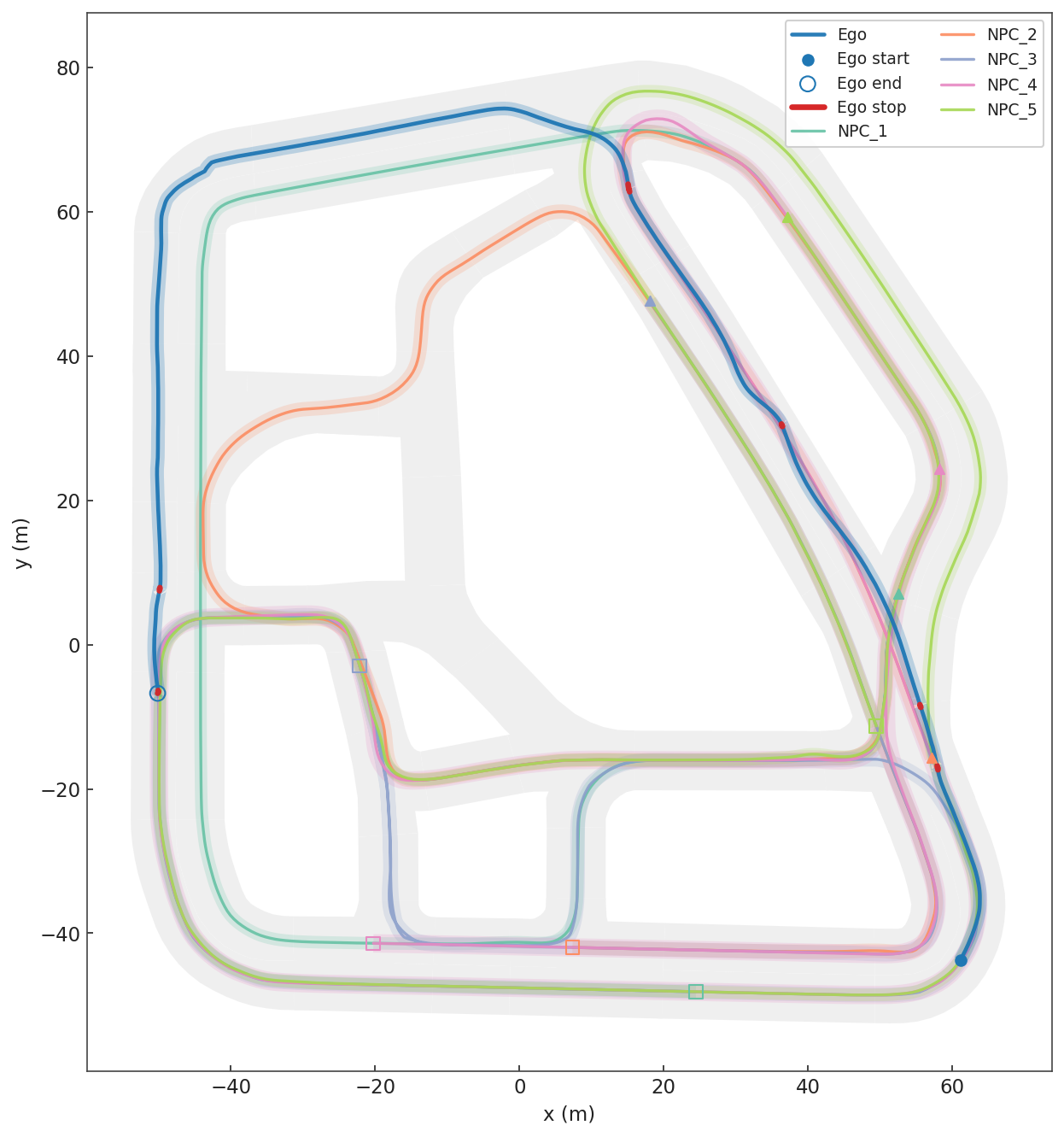}\label{fig:context_b}}
    
    \caption{Context and macroscopic outcome of the multi-agent interaction experiment. Red markers in (b) denote Ego stops.}
    \label{fig:context_and_outcome} 
\end{figure*}

\begin{figure}[!t]
\centering
\includegraphics[width=\columnwidth]{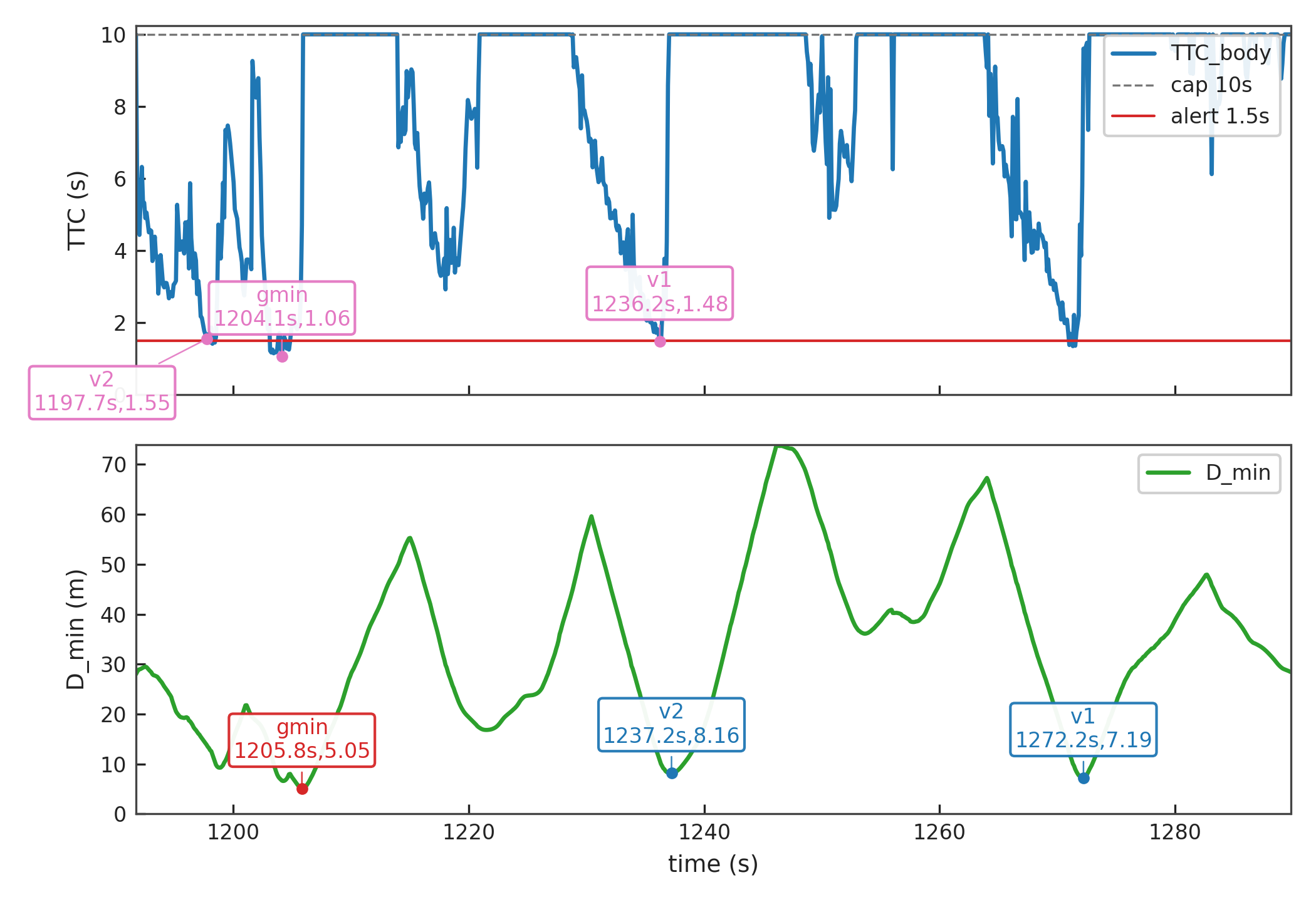} 
\caption{Continuous safety metrics over the full experiment span. 
\textbf{(Top)} Time-to-Collision ($TTC_{\text{body}}$) with a 1.5~s alert threshold (red line). 
\textbf{(Bottom)} Minimum inter-agent distance ($D_{\min}$). Key events such as the global minimum (\textbf{gmin}) and two representative valleys (\textbf{v1}, \textbf{v2}) are annotated for analysis.}
\label{fig:safety_metrics} 
\end{figure}

We evaluate the SUT in a dense urban-intersection task where the Autoware-controlled ego vehicle repeatedly interacts with five NPC vehicles. The experiment spans 98.4~s (from $t=1191.73$ to $1290.13$~s), comprising 985 frames. \textbf{Figure~\ref{fig:context_and_outcome}} summarizes the environmental context and macroscopic trajectories, while \textbf{Figure~\ref{fig:safety_metrics}} presents the continuous safety metrics from this interaction.

To move beyond a binary success/fail verdict, we analyze the interaction's continuous safety profile using our established spatial ($D_{\min}$) and temporal ($TTC_{\text{body}}$) metrics. Our analysis immediately pinpoints the single most critical event of the entire scenario, which occurs around $t \approx 1205$~s. As shown in Fig.~\ref{fig:safety_metrics}, the system recorded its global minimum $TTC_{\text{body}}$ of \textbf{1.06~s} (at $t=1204.1$~s), immediately followed by the global minimum $D_{\min}$ of \textbf{5.05~m} (at $t=1205.8$~s). This tense sequence corresponds to a decisive gap-acceptance maneuver where the ego vehicle threads through dense, crossing traffic.

Beyond this peak-risk event, the framework also quantifies less critical but significant interactions. For instance, the $D_{\min}$ trace reveals other notable valleys at $t\approx1272.2$~s (7.19~m) and $t\approx1237.2$~s (8.16~m), corresponding to yielding for cross-traffic and re-accelerating after clearance, respectively. Similarly, the $TTC_{\text{body}}$ trace shows other transient dips, such as 1.48~s at $t\approx1236.2$~s, which were all successfully resolved by the planner.

In essence, the framework transforms a complex, minute-long interaction into an interpretable safety narrative. It automatically identifies and quantifies the single most critical event while also providing continuous, reproducible evidence for all other maneuvers, thus constructing a comprehensive and evidence-based safety case.

%===========================fig：multi-agent interaction experiment================

%===============fig：safety metrics over the full experiment span===================

\subsection{Summary}
This section demonstrated the three-stage methodology as a single, coherent diagnostic process where each stage logically builds upon the last. The analysis began in \textbf{Stage 1} by identifying an important behavioral characteristic of Autoware, namely its sensitivity to processing latency. In \textbf{Stage 2}, controlled latency injections amplified this observed phenomenon, allowing us to quantify its negative impact up to the boundary of system failure. This process revealed a sharp performance degradation as the total end-to-end latency reached approximately 49 ms, establishing a clear performance boundary against an internal systemic stressor. Finally, \textbf{Stage 3} provided the complementary analysis, stress-testing the SUT against high external environmental complexity. The results from this scenario, run with zero injected latency, showed that while the heavy computational load increased the system's natural operational latency, it remained well within the 49 ms boundary identified in Stage 2, thus providing a direct explanation for its ability to maintain safe performance.

\section{Conclusion and Future Work}
\label{sec:conclusion}

This paper introduced MMRHP, a miniature mixed-reality HIL platform designed to address the critical need for an auditable and reproducible metrology foundation in the closed-loop evaluation of autonomous driving systems. By integrating a unified spatio-temporal measurement baseline with a structured, SOTIF-oriented workflow, we elevate miniature testing from functional demonstration to quantitative, comparable experimentation. Rigorous validation confirmed the platform's centimeter-level spatial accuracy and characterized its decomposable latency profile. A subsequent case study on Autoware demonstrated the methodology's effectiveness, successfully identifying inherent performance signatures and quantifying critical system boundaries, such as a performance cliff induced by a 40 ms latency injection. Ultimately, this work provides a practical and principled path toward credible, audit-ready closed-loop evaluation.

While the proposed framework establishes a robust metrological baseline, a primary limitation is the fidelity of the physical actuator, as the miniature vehicle employs a simplified dynamic model. Future work will directly target this sim-to-real gap through advanced system identification and high-fidelity models. Building on this methodology, further research can extend to new domains, such as studying performance degradation under perception challenges (e.g., sensor noise). The framework can also be adapted for multi-agent testing, enabling the investigation of inherent V2X challenges like distributed clock synchronization and multi-view ground truth fusion.

% \nocite{che2021test,ghiurau2020arcar,baruffa2020mixed,feng2020safety,mitchell2019multi}

% \bibliographystyle{IEEEtran}

\bibliographystyle{IEEEtran}
\bibliography{references}

\begin{thebibliography}{10}
\providecommand{\url}[1]{#1}
\csname url@samestyle\endcsname
\providecommand{\newblock}{\relax}
\providecommand{\bibinfo}[2]{#2}
\providecommand{\BIBentrySTDinterwordspacing}{\spaceskip=0pt\relax}
\providecommand{\BIBentryALTinterwordstretchfactor}{4}
\providecommand{\BIBentryALTinterwordspacing}{\spaceskip=\fontdimen2\font plus
\BIBentryALTinterwordstretchfactor\fontdimen3\font minus \fontdimen4\font\relax}
\providecommand{\BIBforeignlanguage}[2]{{%
\expandafter\ifx\csname l@#1\endcsname\relax
\typeout{** WARNING: IEEEtran.bst: No hyphenation pattern has been}%
\typeout{** loaded for the language `#1'. Using the pattern for}%
\typeout{** the default language instead.}%
\else
\language=\csname l@#1\endcsname
\fi
#2}}
\providecommand{\BIBdecl}{\relax}
\BIBdecl

\bibitem{KALRA2016182}
\BIBentryALTinterwordspacing
N.~Kalra and S.~M. Paddock, ``Driving to safety: How many miles of driving would it take to demonstrate autonomous vehicle reliability?'' \emph{Transportation Research Part A: Policy and Practice}, vol.~94, pp. 182--193, 2016. [Online]. Available: \url{https://www.sciencedirect.com/science/article/pii/S0965856416302129}
\BIBentrySTDinterwordspacing

\bibitem{development_ISO_21448}
A.~Schnellbach and G.~Griessnig, ``Development of the iso 21448,'' in \emph{Systems, Software and Services Process Improvement}, A.~Walker, R.~V. O'Connor, and R.~Messnarz, Eds.\hskip 1em plus 0.5em minus 0.4em\relax Cham: Springer International Publishing, 2019, pp. 585--593.

\bibitem{pegasus2019method}
\BIBentryALTinterwordspacing
{PEGASUS Projekt}, ``Pegasus method,'' German Federal Ministry for Economic Affairs and Energy, Tech. Rep., 2019. [Online]. Available: \url{https://www.pegasusprojekt.de/files/tmpl/Pegasus-Abschlussveranstaltung/PEGASUS-Gesamtmethode.pdf}
\BIBentrySTDinterwordspacing

\bibitem{WOS_000982607400012}
P.~Koopman, ``Ul 4600: What to include in an autonomous vehicle safety case,'' \emph{COMPUTER}, vol.~56, no.~5, pp. 101--104, MAY 2023.

\bibitem{birkemeyer2023scenario}
L.~Birkemeyer, C.~King, and I.~Schaefer, ``Is scenario generation ready for sotif? a systematic literature review,'' in \emph{2023 IEEE 26th International Conference on Intelligent Transportation Systems (ITSC)}.\hskip 1em plus 0.5em minus 0.4em\relax IEEE, 2023, pp. 472--479.

\bibitem{hou2023twin}
Z.~Hou, S.~Wang, H.~Liu, Y.~Yang, and Y.~Zhang, ``Twin scenarios establishment for autonomous vehicle digital twin empowered sotif assessment,'' \emph{IEEE Transactions on Intelligent Vehicles}, vol.~9, no.~1, pp. 1965--1976, 2024.

\bibitem{riedmaier2020survey}
S.~Riedmaier, T.~Ponn, D.~Ludwig, B.~Schick, and F.~Diermeyer, ``Survey on scenario-based safety assessment of automated vehicles,'' \emph{IEEE Access}, vol.~8, pp. 87\,456--87\,477, 2020.

\bibitem{pietruch2020overview}
M.~Pietruch, A.~M{\l}yniec, and A.~Wetula, ``An overview and review of testing methods for the verification and validation of adas, active safety systems, and autonomous driving,'' \emph{Mining--Informatics, Automation and Electrical Engineering}, vol.~58, no.~1, pp. 19--27, 2020.

\bibitem{zhang2024virtual}
\BIBentryALTinterwordspacing
T.~Zhang, H.~Liu, W.~Wang, and X.~Wang, ``Virtual tools for testing autonomous driving: A survey and benchmark of simulators, datasets, and competitions,'' \emph{Electronics}, vol.~13, no.~17, 2024. [Online]. Available: \url{https://www.mdpi.com/2079-9292/13/17/3486}
\BIBentrySTDinterwordspacing

\bibitem{rajabli_svvsac_review}
N.~Rajabli, F.~Flammini, R.~Nardone, and V.~Vittorini, ``Software verification and validation of safe autonomous cars: A systematic literature review,'' \emph{IEEE Access}, vol.~9, pp. 4797--4819, 2021.

\bibitem{zhang2024survey}
H.~Zhang, Y.~Wang, J.~Chen, Z.~Li, H.~Wang, Z.~Zhang, Z.~Wang, Y.~Wang, and Y.-Q. Li, ``A survey of sim-to-real methods in rl: Progress, prospects and challenges with foundation models,'' \emph{arXiv preprint arXiv:2402.13187}, 2024.

\bibitem{liu2024vil}
\BIBentryALTinterwordspacing
Z.~Zhang, G.~Badakis, M.~Galanis, A.~Bavarşi, E.~van Hassel, M.~Alirezaei, and S.~Haesaert, ``A vehicle-in-the-loop simulator with ai-powered digital twins for testing automated driving controllers,'' 2025. [Online]. Available: \url{https://arxiv.org/abs/2507.02313}
\BIBentrySTDinterwordspacing

\bibitem{quinlan2010bringing}
M.~Quinlan, T.-C. Au, J.~Zhu, N.~Stiurca, and P.~Stone, ``Bringing simulation to life: A mixed reality autonomous intersection,'' in \emph{2010 IEEE/RSJ International Conference on Intelligent Robots and Systems}.\hskip 1em plus 0.5em minus 0.4em\relax IEEE, 2010, pp. 6083--6088.

\bibitem{gechter2014towards}
F.~Gechter, B.~Dafflon, P.~Gruer, and A.~Koukam, ``Towards a hybrid real/virtual simulation of autonomous vehicles for critical scenarios,'' in \emph{The Sixth International Conference on Advances in System Simulation (SIMUL 2014)}, 2014, pp. 14--17.

\bibitem{feng2018augmented}
Y.~Feng, C.~Yu, S.~Xu, H.~X. Liu, and H.~Peng, ``An augmented reality environment for connected and automated vehicle testing and evaluation,'' in \emph{2018 IEEE Intelligent Vehicles Symposium (IV)}.\hskip 1em plus 0.5em minus 0.4em\relax IEEE, 2018, pp. 1549--1554.

\bibitem{varga2020mixed}
B.~Varga, M.~Szalai, {\'A}.~Feh{\'e}r, S.~Aradi, and T.~Tettamanti, ``Mixed-reality automotive testing with sensoris,'' \emph{Periodica Polytechnica Transportation Engineering}, vol.~48, no.~4, pp. 357--362, 2020.

\bibitem{kneissl2020mixed}
M.~Kneissl, S.~vom Dorff, A.~Molin, M.~Denniel, T.~D. Son, N.~O. Lleras, H.~Esen, and S.~Hirche, ``Mixed-reality testing of multi-vehicle coordination in an automated valet parking environment,'' \emph{IFAC-PapersOnLine}, vol.~53, no.~2, pp. 17\,564--17\,571, 2020.

\bibitem{szalai2020mixed}
M.~Szalai, B.~Varga, T.~Tettamanti, and V.~Tihanyi, ``Mixed reality test environment for autonomous cars using unity 3d and sumo,'' in \emph{2020 IEEE 18th World Symposium on Applied Machine Intelligence and Informatics (SAMI)}.\hskip 1em plus 0.5em minus 0.4em\relax IEEE, 2020, pp. 73--78.

\bibitem{drechsler2022dynamic}
M.~F. Drechsler, V.~Sharma, F.~Reway, C.~Sch{\"u}tz, and W.~Huber, ``Dynamic vehicle-in-the-loop: A novel method for testing automated driving functions,'' \emph{SAE International Journal of Connected and Automated Vehicles}, vol.~5, no. 12-05-04-0029, pp. 367--380, 2022.

\bibitem{mokhtarian2024survey}
A.~Mokhtarian, J.~Xu, P.~Scheffe, M.~Kloock, S.~Sch{\"a}fer, H.~Bang, V.-A. Le, S.~Ulhas, J.~Betz, S.~Wilson \emph{et~al.}, ``A survey on small-scale testbeds for connected and automated vehicles and robot swarms,'' \emph{arXiv preprint arXiv:2408.14199}, 2024.

\bibitem{li2025autonomous}
D.~Li, P.~Auerbach, and O.~Okhrin, ``Autonomous driving small-scale cars: A survey of recent development,'' \emph{IEEE Transactions on Intelligent Transportation Systems}, 2025.

\bibitem{Tae_sil_mr_agv_ap}
H.~Tae, S.~Yeo, S.~Hwang, S.~Park, and G.~Hwang, ``System-in-the-loop test system with mixed-reality for autonomous ground vehicle (agv) and military applications,'' \emph{IEEE Access}, vol.~13, pp. 46\,383--46\,394, 2025.

\bibitem{verma2021implementation}
A.~Verma, S.~Bagkar, N.~V. Allam, A.~Raman, M.~Schmid, and V.~N. Krovi, ``Implementation and validation of behavior cloning using scaled vehicles,'' \emph{SAE Technical Paper Series}, vol.~1, 2021.

\bibitem{vargas2024design}
D.~Vargas, E.~Haque, M.~Carroll, D.~Perez, T.~Roman, P.~Nguyen, and G.~Habibi, ``Design and implementation of smart infrastructures and connected vehicles in a minicity platform,'' in \emph{2024 IEEE 27th International Conference on Intelligent Transportation Systems (ITSC)}.\hskip 1em plus 0.5em minus 0.4em\relax IEEE, 2024, pp. 513--520.

\bibitem{balaji2019deepracer}
B.~Balaji, S.~Mallya, S.~Genc, S.~Gupta, L.~Dirac, V.~Khare, G.~Roy, T.~Sun, Y.~Tao, B.~Townsend \emph{et~al.}, ``Deepracer: Educational autonomous racing platform for experimentation with sim2real reinforcement learning,'' \emph{arXiv preprint arXiv:1911.01562}, 2019.

\bibitem{bulsara2020obstacle}
A.~Bulsara, A.~Raman, S.~Kamarajugadda, M.~Schmid, and V.~N. Krovi, ``Obstacle avoidance using model predictive control: An implementation and validation study using scaled vehicles,'' \emph{SAE Technical Paper Series}, vol.~1, 2020.

\bibitem{argui2023mixed-reality}
I.~Argui, M.~Gu{\'e}riau, and S.~Ainouz, ``A mixed-reality framework based on depth camera for safety testing of autonomous navigation systems,'' in \emph{2023 IEEE 26th International Conference on Intelligent Transportation Systems (ITSC)}.\hskip 1em plus 0.5em minus 0.4em\relax IEEE, 2023, pp. 2050--2055.

\bibitem{liu2020mobile}
Y.~Liu, G.~Novotny, N.~Smirnov, W.~Morales-Alvarez, and C.~Olaverri-Monreal, ``Mobile delivery robots: Mixed reality-based simulation relying on ros and unity 3d,'' in \emph{2020 IEEE Intelligent Vehicles Symposium (IV)}.\hskip 1em plus 0.5em minus 0.4em\relax IEEE, 2020, pp. 15--20.

\bibitem{chen2009mixed}
I.~Y.-H. Chen, B.~MacDonald, and B.~Wunsche, ``Mixed reality simulation for mobile robots,'' in \emph{2009 IEEE International Conference on Robotics and Automation}.\hskip 1em plus 0.5em minus 0.4em\relax IEEE, 2009, pp. 232--237.

\bibitem{stager2018scaled}
A.~Stager, L.~Bhan, A.~Malikopoulos, and L.~Zhao, ``A scaled smart city for experimental validation of connected and automated vehicles,'' \emph{IFAC-PapersOnLine}, vol.~51, no.~9, pp. 130--135, 2018.

\bibitem{kloock2021cyber}
M.~Kloock, P.~Scheffe, J.~Maczijewski, A.~Kampmann, A.~Mokhtarian, S.~Kowalewski, and B.~Alrifaee, ``Cyber-physical mobility lab: An open-source platform for networked and autonomous vehicles,'' in \emph{2021 European Control Conference (ECC)}.\hskip 1em plus 0.5em minus 0.4em\relax IEEE, 2021, pp. 1937--1944.

\bibitem{tian2024icat}
Z.~Tian, Y.~He, B.~Tian, R.~Zhong, E.~Foorginejad, and W.~Shi, ``Icat: an indoor connected and autonomous testbed for vehicle computing,'' in \emph{2024 IEEE International Conference on Mobility, Operations, Services and Technologies (MOST)}.\hskip 1em plus 0.5em minus 0.4em\relax IEEE, 2024, pp. 242--250.

\bibitem{hyldmar2019fleet}
N.~Hyldmar, Y.~He, and A.~Prorok, ``A fleet of miniature cars for experiments in cooperative driving,'' in \emph{2019 International Conference on Robotics and Automation (ICRA)}.\hskip 1em plus 0.5em minus 0.4em\relax IEEE, 2019, pp. 3238--3244.

\bibitem{dong2023mixed}
J.~Dong, Q.~Xu, J.~Wang, C.~Yang, M.~Cai, C.~Chen, Y.~Liu, J.~Wang, and K.~Li, ``Mixed cloud control testbed: Validating vehicle-road-cloud integration via mixed digital twin,'' \emph{IEEE Transactions on Intelligent Vehicles}, vol.~8, no.~4, pp. 2723--2736, 2023.

\bibitem{abboush2024virtual}
M.~Abboush, C.~Knieke, and A.~Rausch, ``A virtual testing framework for real-time validation of automotive software systems based on hardware in the loop and fault injection,'' \emph{Sensors}, vol.~24, no.~12, p. 3733, 2024.

\bibitem{szalay2021next}
Z.~Szalay, ``Next generation x-in-the-loop validation methodology for automated vehicle systems,'' \emph{IEEE Access}, vol.~9, pp. 35\,616--35\,632, 2021.

\bibitem{scheffe2023scaled}
P.~Scheffe and B.~Alrifaee, ``A scaled experiment platform to study interactions between humans and cavs,'' in \emph{2023 IEEE Intelligent Vehicles Symposium (IV)}.\hskip 1em plus 0.5em minus 0.4em\relax IEEE, 2023, pp. 1--6.

\bibitem{bruggner2021model}
D.~Bruggner, A.~Hegde, F.~S. Acerbo, D.~Gulati, and T.~D. Son, ``Model in the loop testing and validation of embedded autonomous driving algorithms,'' in \emph{2021 IEEE Intelligent Vehicles Symposium (IV)}, 2021, pp. 136--141.

\bibitem{vedder2018low}
B.~Vedder, J.~Vinter, and M.~Jonsson, ``A low-cost model vehicle testbed with accurate positioning for autonomous driving,'' \emph{Journal of Robotics}, vol. 2018, no.~1, p. 4907536, 2018.

\bibitem{merker2023measurement}
S.~Merker, S.~Pastel, D.~B{\"u}rger, A.~Schwadtke, and K.~Witte, ``Measurement accuracy of the htc vive tracker 3.0 compared to vicon system for generating valid positional feedback in virtual reality,'' \emph{Sensors}, vol.~23, no.~17, p. 7371, 2023.

\bibitem{hsiao2022multi}
T.~Hsiao, S.~J. Sheu, and R.~He, ``A multi-precision indoor localization strategy based on hybrid vive and adaptive monte carlo method,'' in \emph{2022 International Automatic Control Conference (CACS)}.\hskip 1em plus 0.5em minus 0.4em\relax IEEE, 2022, pp. 1--6.

\end{thebibliography}
% %%===================================== 

% \end{thebibliography}

% --- Biography Section (Placeholder) ---
% \newpage
\begin{IEEEbiographynophoto}{Mingxin Li}
received the B.S. degree in computer science and technology from China University of Mining and Technology, Beijing, China, in 2021, and the M.S. degree in Information Technology from The Hong Kong Polytechnic University in 2023. He is currently a Research Assistant with the City University of Hong Kong, Hong Kong. His research interests include autonomous driving and robotic manipulation.
\end{IEEEbiographynophoto}

\begin{IEEEbiographynophoto}{Haibo Hu} 
received the M.S. degree from University of Electronic Science and Technology of China in 2019. He worked as a senior development engineer at Alibaba and ByteDance from 2020 to 2023, respectively. He is currently working toward the Ph.D. degree with City University of Hong Kong, Hong Kong, China. His research interests include autonomous driving, large model applications, and AI storage device development.
\end{IEEEbiographynophoto}

\begin{IEEEbiographynophoto}{Jinghuai Deng}
received the B.S. and M.S. degrees in Vehicle Engineering from the School of Transportation Science and Engineering, Beihang University, Beijing, China. He is currently a Ph.D. candidate with the Department of Computer Science, City University of Hong Kong, under the supervision of Prof. Jianping Wang. His research focuses on autonomous driving systems, with particular interests in teleoperation technology and system safety.
\end{IEEEbiographynophoto}

\begin{IEEEbiographynophoto}{Yuchen Xi} 
received the B.S. degree in computer science and technology from Xiamen University Tan Kah Kee College, Zhangzhou, China, in 2022, and the M.S. degree in advanced computer science from Newcastle University, Newcastle upon Tyne, U.K., in 2023. He is currently a Research Assistant with the City University of Hong Kong, Hong Kong, China. His research interests include computer vision, autonomous driving simulation, and machine learning.
\end{IEEEbiographynophoto}

\begin{IEEEbiographynophoto}{Xinhong Chen}
received the B.Eng. degree in software engineering from Sun Yat-Sen University, Guangdong, China, in 2018, and the Ph.D. degree in computer science from City University of Hong Kong, in 2022. He is currently a Research Associate with the Department of Computer Science at City University of Hong Kong, Hong Kong. His research interests including natural language processing, sentiment analysis, autonomous driving, and causality mining.
\end{IEEEbiographynophoto}
\begin{IEEEbiographynophoto}{Jianping Wang}
received the B.S. and M.S. degrees in computer science from Nankai University, Tianjin, China, in 1996 and 1999, respectively, and the Ph.D. degree in computer science from The University of Texas at Dallas, in 2003. She is currently the Dean of College of Computing and a Chair Professor with the Department of Computer Science, City University of Hong Kong. Her research interests include security, autonomous driving, cloud computing, edge computing, and machine learning
\end{IEEEbiographynophoto}

\vfill

\end{document}